\documentclass[journal]{IEEEtran}
\usepackage{times}
\usepackage[pdftex]{graphicx}
\usepackage{subfigure}
\usepackage{amsmath,amssymb,amsopn,amstext,amsfonts}
\usepackage{cancel}
\usepackage[space]{cite}
\usepackage{pdfsync}
\usepackage{balance}
\usepackage{color}
\usepackage{mathtools}
\usepackage{algpseudocode}
\usepackage[ruled,vlined,linesnumbered]{algorithm2e}

\algnewcommand\algorithmicforeach{\textbf{for each}}
\algdef{S}[FOR]{ForEach}[1]{\algorithmicforeach\ #1\ \algorithmicdo}

\usepackage{bm}
\usepackage{subfigure}
\usepackage{diagbox}
\usepackage{epstopdf}
\usepackage{pifont}
\usepackage{fixltx2e}
\usepackage{amsmath}
\usepackage{multirow}
\usepackage{url}
\usepackage{verbatim}
\usepackage{footmisc}
\usepackage{mathabx}
\usepackage{physics}
\usepackage[linkcolor=black,citecolor=black,urlcolor=black,colorlinks=true]{hyperref}
\usepackage{subfigure}
\usepackage{color}
\usepackage{amssymb,mathrsfs,amsmath}
\usepackage[skip=3pt, font={small}]{caption} 
\usepackage{tabularx}
\usepackage{tikz}
\usepackage{amsthm}
\usepackage{wrapfig}
\newcommand{\todo}[1]{\textcolor{red}{\emph{\bf#1}}}

\usepackage{tabulary}
\usepackage{booktabs}
\newtheoremstyle{mystyle}
{}
{}
{\itshape}
{}
{\bfseries}
{.}
{ }
{\thmname{#1}\thmnumber{ #2}\thmnote{ (#3)}}

\theoremstyle{mystyle}

\newcommand{\sqdiamond}[1][fill=black]{\tikz [x=1.2ex,y=1.85ex,line width=.1ex,line join=round, yshift=-0.285ex] \draw  [#1]  (0,.5) -- (.5,1) -- (1,.5) -- (.5,0) -- (0,.5) -- cycle;}%
\newcommand{\MyDiamond}[1][fill=black]{\mathop{\raisebox{-0.275ex}{$\sqdiamond[#1]$}}}

\graphicspath{{figures/}}
\DeclareGraphicsExtensions{.png,.jpg,.eps,.pdf}

\title{\LARGE \bf
	R$^2$LIVE: A Robust, Real-time, LiDAR-Inertial-Visual tightly-coupled state Estimator and mapping
}

\author{
	Jiarong Lin, Chunran Zheng, Wei Xu, and Fu Zhang 
	\thanks{J. Lin, C. Zheng, W. Xu and F. Zhang are with the Department of Mechanical Engineering, Hong Kong University, Hong Kong SAR., China. {\tt\small $\{$jiarong.lin, zhengcr, wuweii,  fuzhang$\}$@hku.hk}
	}
}%

\begin{document}
	\maketitle
	\newcommand{\note}[1]{\textcolor{red}{\emph{\bf#1}}}
	\newcommand\footnoteref[1]{\protected@xdef\@thefnmark{\ref{#1}}\@footnotemark}
	\newlength{\bibitemsep}\setlength{\bibitemsep}{.0238\baselineskip}
	\newlength{\bibparskip}\setlength{\bibparskip}{0pt}
	\let\oldthebibliography\thebibliography
	\renewcommand\thebibliography[1]{%
		\oldthebibliography{#1}%
		\setlength{\parskip}{\bibitemsep}%
		\setlength{\itemsep}{\bibparskip}%
	}
	\begin{abstract}
		In this letter, we propose a robust, real-time tightly-coupled multi-sensor fusion framework, which fuses measurement from LiDAR, inertial sensor, and visual camera to achieve robust and accurate state estimation. Our proposed framework is composed of two parts: the filter-based odometry and factor graph optimization. 
		To guarantee real-time performance, we estimate the state within the framework of error-state iterated Kalman-filter, and further improve the overall precision with our factor graph optimization.
		Taking advantage of measurement from all individual sensors, our algorithm is robust enough to various visual failure, LiDAR-degenerated scenarios, and is able to run in real-time on an on-board computation platform, as shown by extensive experiments conducted in indoor, outdoor, and mixed environment of different scale. Moreover, the results show that our proposed framework can improve the accuracy of state-of-the-art LiDAR-inertial or visual-inertial odometry.
		To share our findings and to make contributions to the community, we open source our codes on our Github\footnote{\url{https://github.com/hku-mars/r2live}\label{foot_github}}.
	\end{abstract}

\section{Introduction}\label{sect_intro}
\begin{figure}[t]
	\centering
	\centering
	\includegraphics[width=1.0\linewidth]{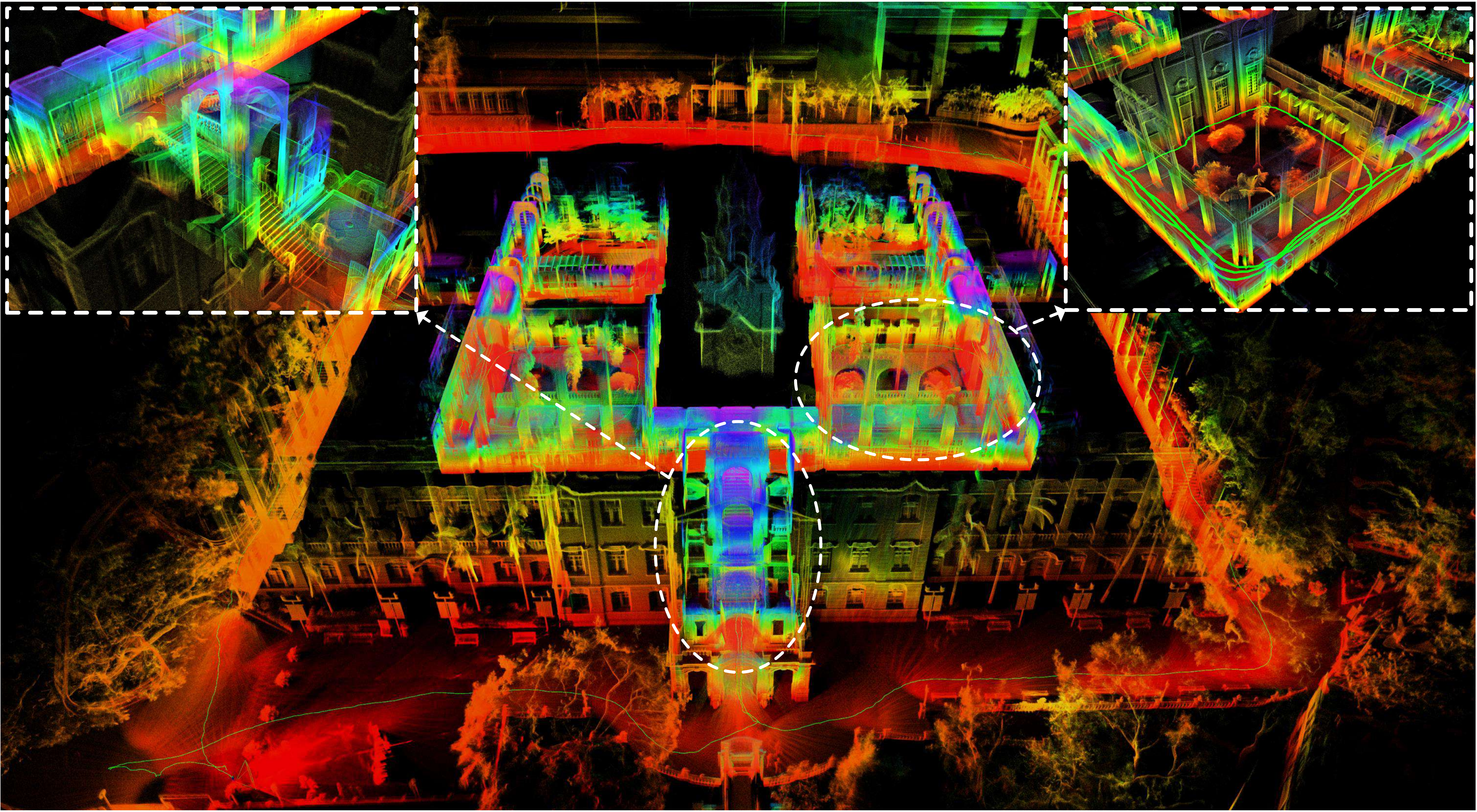}
	\caption{We use our proposed method to reconstruct a high precision, large scale, indoor-outdoor, dense 3D maps of the main building of the University of Hong Kong (HKU). The green path is the computed trajectory and the 3D points are colored by height.}
	\label{fig_cover}
	\vspace{-1.0cm}
\end{figure}

 With the capacity of estimating ego-motion in six degrees of freedom (DOF) and simultaneously building dense and high precision maps of surrounding environments, LiDAR-based SLAM has been widely applied in the field of autonomous driving vehicles \cite{levinson2011towards}, drones \cite{bry2012state, gao2019flying}, and etc. With the development of LiDAR technologies, the emergence of low-cost LiDARs (e.g., Livox LiDAR \cite{liu2020low}) makes LiDAR more accessible. Following this trend, a number of related works \cite{xu2020fast,lin2020decentralized, liu2020balm, lidarcalib} draw the attention of the community to this field of research.
  However, LiDAR-based SLAM methods easily fail (i.e., degenerate) in those scenarios with few available geometry features, which is more critical for those LiDARs with small FoV \cite{lin2020loam}. In this work, to address the degeneration problems of LiDAR-based odometry, we propose a LiDAR-inertial-visual fusion framework to obtain the state estimation of higher robustness and accuracy.  {
 	The main contributions of our work are:
 	\begin{itemize}
	\item We take advantage of measurements from LiDAR, inertial and camera sensors and fuse them in a tight-coupled way. Experiments show that our method is robust enough in various challenging scenarios with aggressive motion, sensor failure, and even in narrow tunnel-like environments with a large number of moving objects and small LiDAR field of view.
	\item We propose a framework with a high-rate filter-based odometry and a low-rarte factor graph optimization. The filter-based odometry fuses the measurements of LiDAR, inertial, and camera sensors within an error-state iterated Kalman filter to achieve real-time performance. The factor graph optimization refines a local map of keyframe poses and visual landmark positions. 
	\item By tightly fusing different types of sensors, we achieve high-accuracy state estimation. Experiment results show that our system is accurate enough to be used to reconstruct large-scale, indoor-outdoor dense 3D maps of building structures (see Fig.~\ref{fig_cover}). 
	\end{itemize}
 }


Our system is carefully engineered and open sourced\footnotemark[1] to benefit the whole robotics community.
	
\section{Related work}\label{sect_intro}
In this section, we review existing works closely related to our work, including LiDAR-only odometry and mapping, LiDAR-Inertial fusion and LiDAR-Inertial-Visual methods.	

\subsection{LiDAR Odometry and Mapping}
Zhang \textit{et al} \cite{zhang2014loam} first proposed a LiDAR odometry and mapping framework, LOAM, that combines ICP method \cite{low2004linear} with point-to-plane and point-to-edge distance. It achieves good odometry and mapping performance by running the two modules at different rates. To make the algorithm run in real time at a computation limited platform, Shan \textit{et al} \cite{shan2018lego} propose a lightweight and ground-optimized LOAM (LeGO-LOAM), which discards unreliable features in the step of ground plane segmentation. 
These works are mainly based on multi-line spinning LiDARs. Our previous work \cite{lin2020loam, lin2019fast} develop an accurate and robust algorithm by considering the low-level physical properties of solid-state LiDARs with small FOV. However, these methods solely based on LiDAR measurements and are very vulnerable to featureless environments or other degenerated scenarios.

\subsection{LiDAR-Inertial Odometry}

The existing works of LiDAR-Inertial fusion can be categorized into two classes: loosely-coupled and the tightly-coupled. Loosely-coupled methods deal with two sensors separately to infer their motion constraints while tightly-coupled approaches directly fuse lidar and inertial measurements through joint-optimization. 
Compared with loosely-coupled methods, tightly-coupled methods show higher robustness and accuracy, therefore drawing increasing research interests recently. For example, authors in \cite{ye2019tightly} propose LIOM which uses a graph optimization based on the priors from LiDAR-Inertial odometry and a rotation-constrained refinement method. Compared with the former algorithm, LIO-SAM \cite{shan2020lio} optimizes a sliding-window of keyframe poses in a factor graph to achieve higher accuracy. Similarly, Li \textit{et al}. propose LiLi-OM \cite{li2020towards} for both conventional and solid-state LiDARs based on sliding window optimization. LINS \cite{qin2020lins} is the first tightly-coupled LIO that solves the 6 DOF ego-motion via iterated Kalman filtering. To lower the high computation load in calculating the Kalman gain, our previous work FAST-LIO \cite{xu2020fast} proposes a new formula of the Kalman gain computation, the resultant computation complexity depends on the state dimension instead of measurement dimension. The work achieves up to 50 Hz odometry and mapping rate while running on embedded computers onboard a UAV.

\subsection{LiDAR-Inertial-Visual Odometry}

On the basis of LiDAR-Inertial methods, LiDAR-Inertial-Visual odometry incorporating measurements from visual sensors shows higher robustness and accuracy. In the work of \cite{zhu2020camvox}, the LiDAR measurements are used to provide depth information for camera images, forming a system similar to RGB-D camera that can leverage existing visual SLAM work such as ORB-SLAM \cite{mur2017orb}. This is a loosely-coupled method as it ignores the direct constraints on state imposed by LiDAR measurements. Zuo \textit{et al} \cite{zuo2019lic} propose a LIC-fusion framework combining IMU measurements, sparse visual features, and LiDAR plane and edge features with online spatial and temporal calibration based on the MSCKF framework, which is claimed more accurate and robust than state-of-the-art methods. In quick succession, their further work termed LIC-Fusion 2.0 \cite{zuo2020lic} refines a novel plane-feature tracking algorithm across multiple LiDAR scans within a sliding-window to make LiDAR scan matching more robust.

To the best of our knowledge, our work is the first open sourced tightly-coupled LiDAR-inertial-visual fusion system. By fusing different types of sensor measurements, we achieve state estimation of higher accuracy and robustness. Extensive results show that our system is more accurate and robust than state-of-the-art LiDAR-inertial and Visual-inertial fusion estimator (e.g., FAST-LIO \cite{xu2020fast}, and VINS-mono\cite{qin2018vins}).

        \section{The overview of our system}
	\begin{figure}[h]
		\centering
		{
			\includegraphics[width=1.0\linewidth]{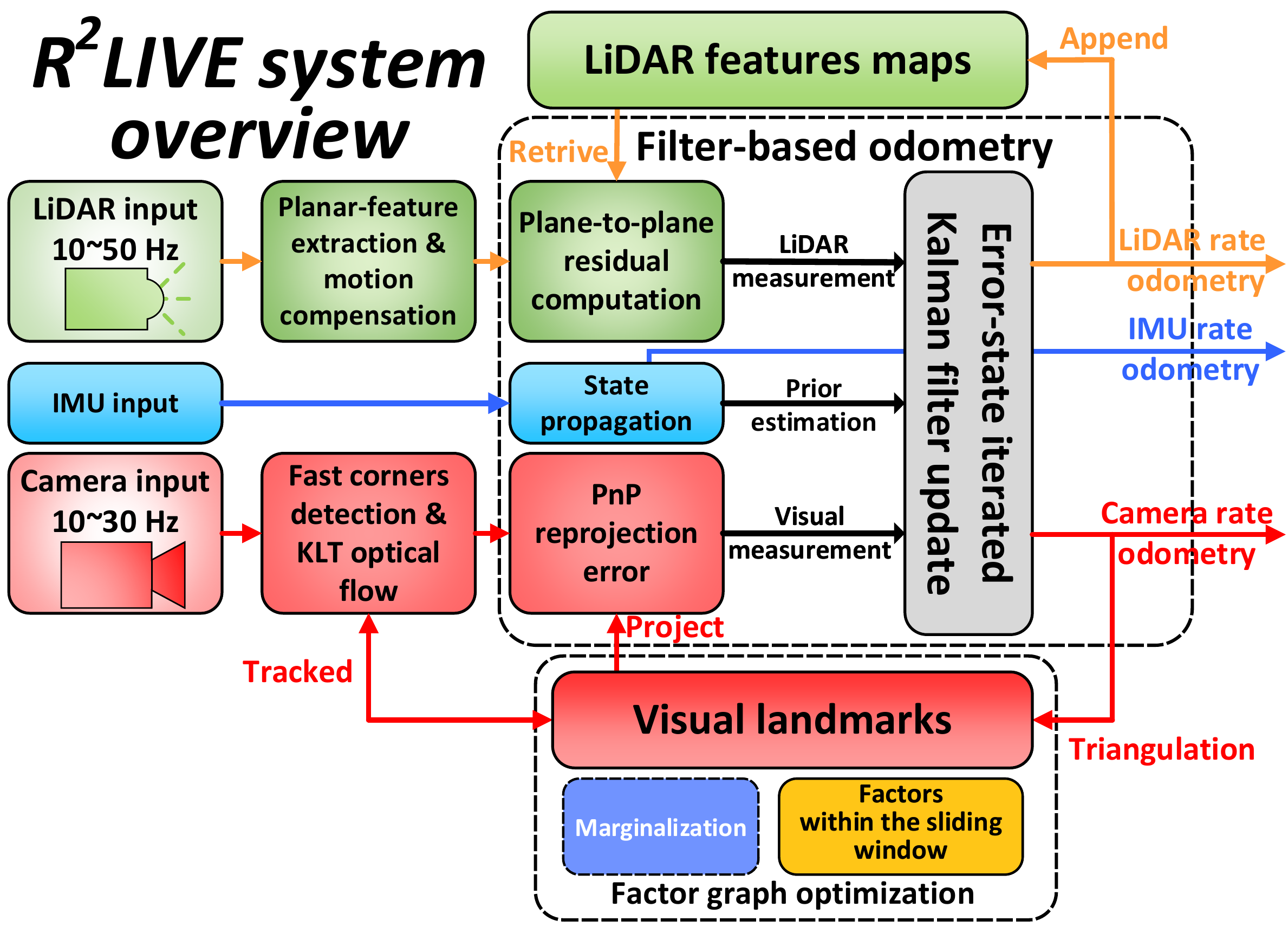}
			\caption{The overview of our proposed method.}
			\label{fig_overview}
			\vspace{-0.8cm}
		}
	\end{figure}

	The overview of our system is shown in Fig.~\ref{fig_overview}, where the filter-based odometry taking advantage of the measurements from LiDAR, camera and inertial sensor, estimates the state within the framework of error-state iterated Kalman filter as detailed in Section \ref{sec:Odom}. To further improve the visual measurements, we leverage the factor graph optimization to refine the visual landmarks within a local sliding window as detailed in Section \ref{sec:fact_graph}.  
	

    	\section{Filter-based odometry}\label{sec:Odom}

\subsection{The boxplus ``$\boxplus$'' and boxminus ``$\boxminus$'' operator}
    In this paper, we make use of the ``$\boxplus$" and ``$\boxminus$" operations encapsulated on a manifold $\mathcal{M}$ to simplify the notations and derivations.
    Let $\mathcal{M}$ be the manifold on which the system state lies. Since a manifold is locally homeomorphic to $\mathbb{R}^n$, where $n$ is the dimension of $\mathcal{M}$, we can use two operators, ``$\boxplus$'' and ``$\boxminus$'', establishing a bijective map between the local neighborhood of $\mathcal{M}$ and its tangent space $\mathbb{R}^n$ \cite{hertzberg2013integrating}:
	\begin{align}
		\boxplus: \mathcal{M}\times \mathbb{R}^n\rightarrow \mathcal{M}, ~~\boxminus: \mathcal{M}\times\mathcal{M} \rightarrow \mathbb{R}^{n}
	\end{align}

	For the compound manifold $\mathcal{M} = SO(3)\times \mathbb{R}^n$, we have:
	\begin{small}
	\begin{align}
		\begin{bmatrix}
			\mathbf{R} \\
			\mathbf{a}_1
		\end{bmatrix} 
		\boxplus 
		\begin{bmatrix}
			\mathbf{r} \\
			\mathbf{a}_2
		\end{bmatrix} 
		\triangleq \begin{bmatrix}
			\mathbf{R}\cdot\mathtt{Exp}(\mathbf{r}) \\
			\mathbf{a}_1 + \mathbf{a}_2
		\end{bmatrix}, \hspace{0.1cm}
		\begin{bmatrix}
			\mathbf{R}_1 \\
			\mathbf{a}_1
		\end{bmatrix} 
		\boxminus
		\begin{bmatrix}
			\mathbf{R}_2 \\
			\mathbf{a}_2
		\end{bmatrix} 
		\triangleq \begin{bmatrix}
			\mathtt{Log}(\mathbf{R}_2 ^T \mathbf{R}_1 ) \\
			\mathbf{a}_1 - \mathbf{a}_2
		\end{bmatrix}\nonumber
	\end{align}
	\end{small}
	where $\mathbf{r}\in \mathbb{R}^3$, $\mathbf{a}_1, \mathbf{a}_2 \in \mathbb{R}^n$, $\mathtt{Exp}(\cdot)$ and $\mathtt{Log}(\cdot)$ denote the Rodrigues' transformation between the rotation matrix and rotation vector\footnote{\url{https://en.wikipedia.org/wiki/Rodrigues\%27_rotation_formula}}.
	
	\subsection{Continuous-time kinematic model}\label{sect_continuous_model}
\begin{figure}[t]
		\centering
		\includegraphics[width=1.0\linewidth]{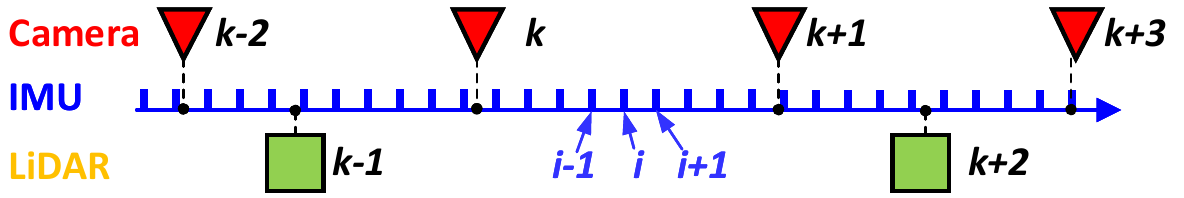}
		\caption{Illustration of the input data sequences, where the frame rate of IMU, camera, and LiDAR is 200 Hz, 20 Hz and 10 Hz, respectively. The notation $i$ denotes the index of IMU data while $k$ denotes the index of LiDAR or camera measurements.}
		\label{fig_datastream}
		\vspace{-0.7cm}
	\end{figure}

		In our current work, we assume that the time offset among all the three sensors, LiDAR, camera, and IMU, are pre-calibrated. Furthmore, we assume the extrinsic between LiDAR and IMU are known as they are usually integrated and calibrated in factory, but estimate the camera IMU extrinsic online. Moreover, a LiDAR typically scans points sequentially, and points in a LiDAR frame could be measured at different body pose. This motion is compensated by IMU back propagation as shown in \cite{xu2020fast}, hence points in a LiDAR frame are assumed to be measured at the same time. With these, the input data sequences of our system can be simplified into Fig. \ref{fig_datastream}.
	
 We assume that the IMU, LiDAR, and camera sensors are rigidly attached together with the extrinsic between LiDAR and IMU (LiDAR frame w.r.t. IMU frame)  as $^I\mathbf{T}_L = (^I\mathbf{R}_L, {^I\mathbf{p}_L})$, and the extrinsic between camera and IMU (camera frame w.r.t. IMU frame) is $^I\mathbf{T}_C = (^I\mathbf{R}_C, {^I\mathbf{p}_C} )$. For the sake of convenience, we take IMU as the body frame, which leads to the following continuous kinematic model:
	\begin{align}
		{^G\dot{\mathbf{p}}_I} = 	{^G{\mathbf{v}}_I}, {^G\dot{\mathbf{v}}_I} = {^G{\mathbf{R}}_I}(\mathbf{a}_m &- \mathbf{b}_a - \mathbf{n}_a)+ {^G{\mathbf{g}}}, \nonumber\\
		\hspace{-0.2cm} {^G{\dot{\mathbf{R}}}_I} = {^G{{\mathbf{R}}}_I} \left[\boldsymbol{\omega}_m-\mathbf{b}_{\mathbf{g}}  - 
		 \mathbf{n}_{\mathbf{g}} \right]_\times &,~  \dot{\mathbf{b}}_{\mathbf{g}}= \mathbf{n}_{\mathbf{bg}}, \dot{\mathbf{b}} = \mathbf{n}_{\mathbf{ba}}
		 \label{eq:cont_model}
	\end{align}
	where ${^G(\cdot)}$ denotes a vector represented in the global frame (i.e. the first gravitational aligned IMU frame \cite{qin2018vins}), $^G\mathbf{R}_I$ and $^G\mathbf{p}_I$ are the attitude and position of the IMU relative to the global frame,  ${^G\mathbf{g}}$ is the gravitational acceleration, $\boldsymbol{\omega}_m$ and $\mathbf{a}_m$ are the raw gyroscope and accelerometer readings, $\mathbf{n}_\mathbf{a}$ and $\mathbf{n}_\mathbf{g}$ are the white noise of IMU measurement, $\mathbf{b}_\mathbf{a}$ and $\mathbf{b}_\mathbf{g}$ are the bias of gyroscope and accelerometer, which are modelled as random walk driven by Gaussian noise $\mathbf{n}_\mathbf{bg}$ and $\mathbf{n}_\mathbf{ba}$.

    \subsection{Discrete IMU model }
    We discretize the continuous model (\ref{eq:cont_model}) at the IMU rate. Let $\mathbf{x}_i$ be the state vector at the $i$-th IMU measurement:
    \begin{align}
    \mathbf{x}_i &=
    \begin{bmatrix}
    ^G\mathbf{R}_{I_i}^T & ^G\mathbf{p}_{I_i}^T & ^{I}\mathbf{R}_{C_i}^T & ^{I}\mathbf{p}_{C_i}^T & ^G\mathbf{v}_{i}^T & \mathbf{b}_{\mathbf{g}_i}^T & \mathbf{b}_{\mathbf{a}_i}^T
    \end{bmatrix}^T \nonumber
	\end{align}
	Discretizing (\ref{eq:cont_model}) by zero-order holder (i.e.,  IMU measurements over one sampling time period $\Delta t$ are constant), we obtain
\begin{equation}
    \mathbf{x}_{t+1} = \mathbf{x}_{i} \boxplus \left(\Delta t\mathbf{f}\left(\mathbf{x}_i, \mathbf{u}_i, \mathbf{w}_i \right)\right)
		\label{eq_true_state_propagate}
\end{equation}
	where
	\begin{align}
    \mathbf{u}_i &=
    \begin{bmatrix}
       \boldsymbol{\omega}_{m_i}^T  & \mathbf{a}_{m_i}^T
    \end{bmatrix}^T, \hspace{0.2cm}
    \mathbf{w}_i =
        \begin{bmatrix}
           \mathbf{n}_{\mathbf{g}_i}^T  & \mathbf{n}_{\mathbf{a}_i}^T &
           \mathbf{n}_{\mathbf{b}\mathbf{g}_i}^T  & \mathbf{n}_{\mathbf{b}\mathbf{a}_i}^T
        \end{bmatrix}^T \hspace{0.2cm} \nonumber  \\
  	 \mathbf{f}&(\mathbf{x}_i, \mathbf{u}_i, \mathbf{w}_i ) = 
  	\begin{bmatrix}
  		\boldsymbol{\omega}_{m_i} - \mathbf{b}_{\mathbf{g}_i} - \mathbf{n}_{\mathbf{g}_i}  \\
  		^G\mathbf{v}_{i} \\
  		\mathbf{0}_{3\times 1} \\
  		\mathbf{0}_{3\times 1} \\
  		^G\mathbf{R}_{I_i}\left( \mathbf{a}_{m_i} - \mathbf{b}_{\mathbf{a}_i} - \mathbf{n}_{\mathbf{g}_i}\right) - {^{G}\mathbf{g}}  \\
  		\mathbf{b}_{\mathbf{g}_i}\\
  		\mathbf{b}_{\mathbf{a}_i}
  	\end{bmatrix} \nonumber
  \end{align}

    \subsection{Propagation}\label{eq:Propagation}
    
	In our work, we leverage an on-manifold iterated error state Kalman filter \cite{he2021embedding} to estimate the state vector $\mathbf x_i$, in which the state estimation error $\delta \hat{\mathbf{x}}_i$ is characterized in the tangent space of the state estimate $\hat{\mathbf{x}}_{i}$: 
	\vspace{-0.3cm}
  \begin{align}
   &\delta \hat{\mathbf{x}}_i \triangleq \mathbf{x}_i \boxminus \hat{\mathbf{x}}_i \nonumber \\
    & = 
   \begin{bmatrix}
    ^G\delta\hat{\mathbf{r}}_{I_i}^T & ^G\delta\hat{\mathbf{p}}_{I_i}^T & ^{{I}}\delta\hat{\mathbf{r}}_{C_i}^T & ^{{I}}\delta\hat{\mathbf{p}}_{C_i}^T & ^G\delta\hat{\mathbf{v}}_{i}^T & \delta\hat{\mathbf{b}}_{\mathbf{g}_i}^T & \delta\hat{\mathbf{b}}_{\mathbf{a}_i}^T
   \end{bmatrix}^T \nonumber \\
   & \sim  \mathcal{N}(\mathbf{0}_{21\times 1}, \boldsymbol{\Sigma}_{\delta \hat{\mathbf{x}}_i})  \label{eq_def_delta_x_i}
   \end{align}
   Note that $\delta \hat{\mathbf{x}}_i \in \mathbb{R}^{21}$ is in minimum dimension (the system dimension 21) and is a random vector with covariance $\boldsymbol{\Sigma}_{\delta \hat{\mathbf{x}}_i}$. $^G\delta\hat{\mathbf{r}}_{I_i}^T$ and ${^{{I}}\delta\hat{\mathbf{r}}_{C_i}^T }$ are:
    $$^G\delta\hat{\mathbf{r}}_{I_i} = \mathtt{Log}( {^G\hat{\mathbf{R}}_{I_i}^T} {^G{\mathbf{R}}_{I_i}^T} )  ,~~ {{^{{I}}\delta\hat{\mathbf{r}}_{C_i}}} = \mathtt{Log}( {^{I}\hat{\mathbf{R}}_{C_i}^T} {^{I}{\mathbf{R}_{C_i}}} )  $$
	
	Once receiving a new IMU measurement, the state estimate is propagated by setting the process noise in (\ref{eq_true_state_propagate}) to zero: 
    \begin{align}
    \hat{\mathbf{x}}_{i+1} = \hat{\mathbf{x}}_{i} \boxplus \left( \Delta t \cdot \mathbf{f}(\hat{\mathbf{x}}_i, \mathbf{u}_i, \mathbf{0}) \right). 
    \label{eq:state_prop}
    \end{align}

    The associated estimation error is propagated in the linearized error space as follows (see \cite{he2021embedding} for more details): 
    \begin{align}
        &\delta{\hat{\mathbf{x}}}_{i+1} = {\mathbf{x}}_{i+1} \boxminus \hat{\mathbf{x}}_{i+1}  \label{eq_delta_x_k_plus_1} \\
        &= \Large(\mathbf{x}_{i} \boxplus \left( \Delta t \cdot \mathbf{f}({\mathbf{x}}_i, \mathbf{u}_i, \mathbf{w}_i) \right)  \Large) \boxminus \left( \hat{\mathbf{x}}_{i}\boxplus
         \left( \Delta t \cdot \mathbf{f}(\hat{\mathbf{x}}_i, \mathbf{u}_i, \mathbf{0}) \right)\right) \nonumber\\
        & \sim \mathcal{N}(\mathbf 0_{21 \times 1},  \boldsymbol{\Sigma}_{\delta \hat{\mathbf{x}}_{i+1}}  ) \nonumber
   \end{align}
   where:
   \begin{small}
   \begin{align}
            & \boldsymbol{\Sigma}_{\delta \hat{\mathbf{x}}_{i+1}} = \mathbf{F}_{\delta{\hat{\mathbf{x}}}} \boldsymbol{\Sigma}_{\delta \hat{\mathbf{x}}_{i}} \mathbf{F}_{\delta{\hat{\mathbf{x}}}}^T + \mathbf{F}_{\mathbf{w}} \mathbf{Q} \mathbf{F}_{\mathbf{w}}^T 
            \label{eq:cov_prop} \\
			&\mathbf{F}_{\delta{\hat{\mathbf{x}}}} = \left. \dfrac{ \partial \left( \delta{\hat{\mathbf{x}}}_{i+1} \right)  }{\partial\delta{\hat{\mathbf{x}}_i}} \right|_{\delta{\hat{\mathbf{x}}_i} = \mathbf{0}, \mathbf{w}_i = \mathbf{0}} , ~~
		 \mathbf{F}_{{\mathbf{w}}} = \left. \dfrac{ \partial \left( \delta{\hat{\mathbf{x}}}_{i+1} \right)  }{\partial{\mathbf{w}_i}} \right|_{\delta{\hat{\mathbf{x}}_i} = \mathbf{0}, \mathbf{w}_i = \mathbf{0}} \nonumber 
   \end{align}
   \end{small}
	with their exact values computed in Appendix. \ref{sect_app_fx_fw}.
	
	The two propagation in (\ref{eq:state_prop}) and (\ref{eq:cov_prop}) starts from the optimal state and covariance estimate after fusing the most recent LiDAR/camera measurement (e.g., the $k$-th measurement, see Section \ref{sec:update}), and repeat until receiving the next LiDAR/camera measurement (e.g., the $(k+1)$-th measurement). The relation between time index $i$ and $k$ is shown in Fig. \ref{fig_datastream}. 

    \subsection{The prior distribution}
   Let the two propagation in (\ref{eq:state_prop}) and (\ref{eq:cov_prop}) stop at the $(k+1)$-th LiDAR/camera measurement (see Fig. \ref{fig_esikf}), and the propagated state estimate and covariance are $ \hat{\mathbf{x}}_{k+1}$ and $\boldsymbol{\Sigma}_{\delta \hat{\mathbf{x}}_{k+1}}$, respectively. They essentially impose a prior distribution for the state $\mathbf x_{k+1}$ before fusing the $(k+1)$-th measurement as below: 
   	\begin{figure}[t]
   	\centering
   	\includegraphics[width=0.8\linewidth]{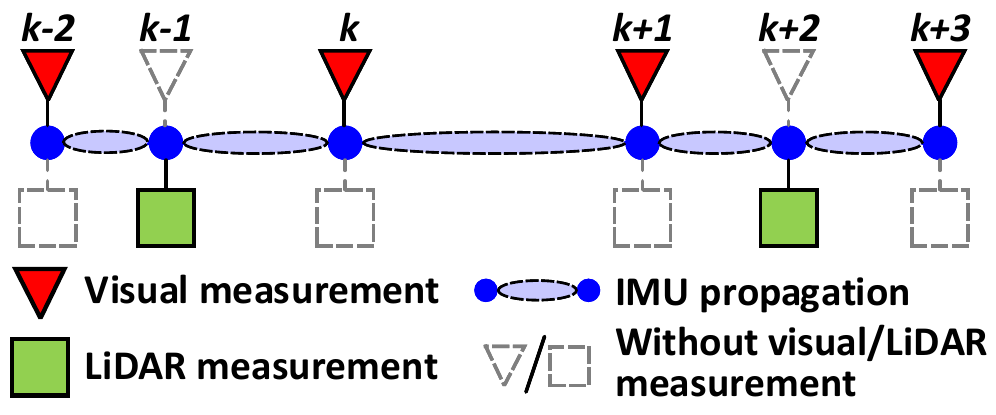}
   	\caption{The illustraction of the update of our error-state iterated Kalman filter.}
   	\label{fig_esikf}
   	\vspace{-1.0cm}
   \end{figure}
   \begin{align}
        \mathbf{x}_{k+1} \boxminus \hat{\mathbf{x}}_{k+1} 
        &\sim \mathcal{N}(\mathbf{0}, \boldsymbol{\Sigma}_{\delta \hat{\mathbf{x}}_{k+1}} ). \label{eq:prior_dist}
   \end{align}
   
   \subsection{Initialization of iterated update}
   
    The prior distribution in (\ref{eq:prior_dist}) will be fused with the LiDAR or camera measurements to produce a maximum a-posterior (MAP) estimate (denoted as $\check{\mathbf x}_{k+1}$) of $\mathbf x_{k+1}$. The  MAP estimate $\check{\mathbf x}_{k+1}$ is initialized as the prior estimate $\hat{\mathbf x}_{k+1}$ and is refined iteratively due to the nonlinear nature of the problem.  In each iteration, the error $\delta{\check{\mathbf{x}}}_{k+1}$ between the true state $\mathbf x_{k+1}$ and the current estimate $\check{\mathbf x}_{k+1}$, defined as 
    \begin{equation}
        \label{eq:error}
        \delta{\check{\mathbf{x}}}_{k+1} \triangleq \mathbf x_{k+1} \boxminus \check{\mathbf x}_{k+1},
    \end{equation}
    will be solved by minimizing the posterior distribution considering the prior in (\ref{eq:prior_dist}) and LiDAR/visual measurements. Therefore, the prior distribution in terms of $\mathbf x_{k+1}$ represented by (\ref{eq:prior_dist}) should be transformed to an equivalent prior distribution in terms of $\delta{\check{\mathbf{x}}}_{k+1}$: 
    \begin{equation}
        \label{eq:trans}
        \begin{split}
            \mathbf{x}_{k+1} \boxminus \hat{\mathbf{x}}_{k+1} &= \left( \check{\mathbf x}_{k+1} \boxplus \delta{\check{\mathbf{x}}}_{k+1} \right) \boxminus \hat{\mathbf{x}}_{k+1} \\
            &\approx \check{\mathbf{x}}_{k+1} \boxminus \hat{\mathbf{x}}_{k+1}  + \boldsymbol{\mathcal{H}} \delta{\check{\mathbf{x}}}_{k+1} \\
            & \sim \mathcal{N}(\mathbf{0}, \boldsymbol{\Sigma}_{\delta \hat{\mathbf{x}}_{k+1}} ),
        \end{split}
    \end{equation}
    where $$\boldsymbol{\mathcal{H}} =  \dfrac{ \left( \check{\mathbf{x}}_{k+1} \boxplus \delta \check{\mathbf{x}}_{k+1}  \right) \boxminus \hat{\mathbf{x}}_{k+1} }{\partial  \delta \check{\mathbf{x}}_{k+1}} |_{ \delta \check{\mathbf{x}}_{k+1}  = \mathbf{0}}$$ is computed in detail in Appendix. \ref{sect_app_Hx}, (\ref{eq:trans}) essentially imposes a prior distribution to $\delta{\check{\mathbf{x}}}_{k+1}$ as below:
    \begin{equation}
        \hspace{-0.2cm}\delta{\check{\mathbf{x}}}_{k+1} \sim \mathcal{N}(-\boldsymbol{\mathcal{H}}^{-1} \left( \check{\mathbf{x}}_{k+1} \boxminus \hat{\mathbf{x}}_{k+1} \right), \boldsymbol{\mathcal{H}}^{-1}  \boldsymbol{\Sigma}_{\delta \hat{\mathbf{x}}_{k+1}} \boldsymbol{\mathcal{H}}^{-T})
        \label{eq:prior_delta}
    \end{equation}

   \subsection{LiDAR measurement}\label{sec:lidar_meas}
      If the $(k+1)$-th measurement is a LiDAR frame, we extract planar feature points from the raw 3D points as in \cite{xu2020fast} and compensate the in-frame motion as in Section \ref{sect_continuous_model}. Denote  ${\boldsymbol{\mathcal{L}}}_{k+1}$ the set of feature points after motion compensation, we compute the residual of each feature point ${^{L}\mathbf{p}_j} \in {\boldsymbol{\mathcal{L}}}_{k+1}$ {where $j$ is the index of feature point} and the superscript $L$ denotes that the point is represented in the LiDAR-reference frame.
      
      With $\check{\mathbf x}_{k+1}$ being the current estimate of ${\mathbf x}_{k+1}$, we can transform ${^{L}\mathbf{p}_j}$ from LiDAR frame to the global frame
      $
      	{^{G}\mathbf{p}_j} = {^G{\check{\mathbf{R}}}_{I_{k+1}}}( ^I\mathbf{R}_{L}{^{L}{\mathbf{p}}_j}  + {^{I}\mathbf{p}_L}) +  {^G\check{\mathbf{p}}_{I_{k+1}}}.
      $
      As the previous LOAM pipeline does in \cite{lin2020loam, xu2020fast}, we search for the nearest planar feature points in the map and use them to fit a plane with normal $\mathbf{u}_j$ and an in-plane point ${{\mathbf{q}}_j }$, the measurement residual is:
      \begin{align}
      	\mathbf{r}_l(\check{\mathbf{x}}_{k+1}, {^{L}{\mathbf{p}}}_{j})=& \mathbf{u}_j^T\left({^{G}\mathbf{p}_j} - {{\mathbf{q}}_j } \right) \label{eq_def_rl_x_kplus1}      
      \end{align}
     Let $\mathbf n_j$ be the measurement noise of the point ${^{L}\mathbf{p}_j}$, we can obtain the true point location ${^{L}\mathbf{p}^{\mathtt{gt}}_j}$ by compensating the noise from ${^{L}\mathbf{p}_j}$:
	  \begin{align}
			{^{L}\mathbf{p}_j} = {^{L}\mathbf{p}^{\mathtt{gt}}_j} + {\mathbf{n}_j},  {\mathbf{n}_j} \sim \mathcal{N}(\mathbf{0}, \boldsymbol{\Sigma}_{\mathbf{n}_j}). 
	  \end{align}
	  This true point location together with the true state $\mathbf x_{k+1}$ should lead to zero residual in (\ref{eq_def_rl_x_kplus1}), i.e.,
	  \begin{small}
      \begin{align}
    	\mathbf 0 = \mathbf{r}_l(\mathbf{x}_{k+1}, {^{L}\mathbf{p}}^{\mathtt{gt}}_{j}) &= \mathbf{r}_l(\check{\mathbf{x}}_{k+1}, {^{L}{\mathbf{p}}}_{j}) + \mathbf{H}^{{l}}_{j} \delta \check{\mathbf{x}}_{k+1} + \boldsymbol{\alpha}_{j}, \label{eq_lidar_meas}
      \end{align}
  	  \end{small}
      which constitutes a posteriori distribution for $\delta \check{\mathbf{x}}_{k+1}$.  In (\ref{eq_lidar_meas}), $\mathbf x_{k+1}$ is parameterized by its error $\delta \check{\mathbf{x}}_{k+1}$ defined in (\ref{eq:error}) and $\boldsymbol{\alpha}_{j} \sim \mathcal{N}(\mathbf{0}, \boldsymbol{\Sigma}_{\boldsymbol{\alpha}_{j}})$:
      \begin{align}
    	\mathbf{H}^{{l}}_{j}  =& \dfrac{ \partial  \mathbf{r}_l(\check{\mathbf{x}}_{k+1} \boxplus \delta{\check{\mathbf{x}}}_{k+1}, {^{L}{\mathbf{p}}}_{j}) }{\partial  \delta{\check{\mathbf{x}}}_{k+1}} |_{\delta{\check{\mathbf{x}}}_{k+1} = \mathbf{0}} \nonumber \\
    	\boldsymbol{\Sigma}_{\boldsymbol{\alpha}_{j}} = & {\mathbf{F}_{{{\mathbf{p}}}_{j}}}  \boldsymbol{\Sigma}_{\mathbf{n}_j}  {\mathbf{F}_{{{\mathbf{p}}}_{j}}^T} \label{eq_def_HL_gamma_alpha}  \\
    	{\mathbf{F}_{{{\mathbf{p}}}_{j}}} =&   \left( \dfrac{ \partial  \mathbf{r}_l(\check{\mathbf{x}}_{k+1}, {^{L}{\mathbf{p}}}_{j}) }{\partial  {^{L}{\mathbf{p}}}_{j} }\right) =  {^G{\check{\mathbf{R}}}_{I_{k+1}}} ^I\mathbf{R}_{L} \nonumber
      \end{align}
  	The detailed computation of $\mathbf{H}^{{l}}_{j}$ can be found in Appendix. \ref{sect_app_HL_alpha}.
    \subsection{Visual measurement}\label{sect_visual_mes}
  	If the $(k+1)$-th frame is a camera image, we extract the FAST corner feature points $\boldsymbol{\mathcal{C}}_{k+1}$ from the undistorted image and use KLT optical flow to track feature points in $\boldsymbol{\mathcal{C}}_{k+1}$ seen by keyframes in the current sliding window (Section \ref{sec:fact_graph}). If a feature point in $\boldsymbol{\mathcal{C}}_{k+1}$ is lost or has not been yet tracked before, we triangulate the new feature point in 3D space (visual landmarks) with the optimal estimated camera poses.
  	
  	The reprojection errors between visual landmarks and its tracked feature points in the $(k+1)$-th frame are used for updating the current state estimate $\check{\mathbf{x}}_{k+1}$. For an extracted corner point ${^{C}\mathbf{p}_{s}} = \begin{bmatrix}
 		u_s & v_s
 	\end{bmatrix}^T \in \boldsymbol{\mathcal{C}}_{k+1}$ {where $s$ is the index of corner point}, its correspondence landmark in 3D space is denoted as $^G\mathbf{P}_{s}$, then the measurement residual of ${^{C}}\mathbf{p}_{s}$ is:
 \begin{equation}
 	\begin{split}
 	& {}^C \mathbf  P_s  = 	\left({^G\check{\mathbf{R}}_{I_{k+1}}}{^I\check{\mathbf{R}}_{C_{k+1}}} \right)^T  {{^G}{\mathbf{P}}_s} \\
 	& \quad \quad \quad \quad \quad \quad \quad \quad - \left({^I\check{\mathbf{R}}_{C_{k+1}}}\right)^T  {{^G}\check{\mathbf{p}}_{I_{k+1}}} -{^I\check{\mathbf{p}}_{C_{k+1}}} \label{eq_visual_residual} \\
 		&\mathbf{r}_c\left({\check{\mathbf{x}}_{k+1}, {^{C}{\mathbf{p}}}_{s}}, {^G\mathbf{P}_{s}}\right) = 
 		{^{C}}\mathbf{p}_{s} - \boldsymbol{\pi}( {^C\mathbf{P}_{s}} )
 	\end{split}
\end{equation} 
where $\boldsymbol{\pi}(\cdot)$ is the pin-hole projection model. 

Now considering the measurement noise, we have:
\begin{align}
	{^G\mathbf{P}_{s}}	= {^G\mathbf{P}_{s}^{\mathtt{gt}}}  + \mathbf{n}_{\mathbf{P}_{s}},~& \mathbf{n}_{\mathbf{P}_{s}} \sim \mathcal{N}(\mathbf{0}, \boldsymbol{\Sigma}_{\mathbf{n}_{\mathbf{P}_{s}}})  \\
	{^{C}\mathbf{p}_{s}} = {^{C}\mathbf{p}_{s}^{\mathtt{gt}}} + \mathbf{n}_{\mathbf{p}_{s}},~& \mathbf{n}_{\mathbf{p}_{s}} \sim \mathcal{N}(\mathbf{0}, \boldsymbol{\Sigma}_{\mathbf{n}_{\mathbf{p}_{s}}}) 
\end{align}
where $ {^G\mathbf{P}_{s}^{\mathtt{gt}}}$ and ${^{C}\mathbf{p}_{s}^{\mathtt{gt}}}$ are the true value of ${^G\mathbf{P}_{s}}$ and ${^{C}\mathbf{p}_{s}} $, respectively. With these, we obtain the first order Taylor expansion of the true zero residual $\mathbf{r}_c(\mathbf{x}_{k+1}, {^{C}\mathbf{p}^{\mathtt{gt}}_{s}})$ as:
\begin{equation}
    \begin{split}
        \mathbf 0 &= \mathbf{r}_c(\mathbf{x}_{k+1}, {^{C}\mathbf{p}}^{\mathtt{gt}}_{s}, {^G\mathbf{P}^{\mathtt{gt}}_{s}}) \\
	& \approx \mathbf{r}_c\left({\check{\mathbf{x}}_{k+1}, {^{C}{\mathbf{p}}}_{s}}, {^G\mathbf{P}_{s}}\right) + \mathbf{H}^{{c}}_{s} \delta \check{\mathbf{x}}_{k+1} + \boldsymbol{\beta}_{s}, \label{eq_visual_meas}
    \end{split}
\end{equation}
which constitutes another posteriori distribution for $\delta \check{\mathbf{x}}_{k+1}$. In (\ref{eq_visual_meas}), $\boldsymbol{\beta}_{s} \sim \mathcal{N}(\mathbf{0}, {\boldsymbol{\Sigma}_{\boldsymbol{\beta}_{s}}}  )$ and:
\begin{align}
	\hspace{-0.2cm}\mathbf{H}^{{c}}_{s}  &=\dfrac{ \partial  \mathbf{r}_c(\check{\mathbf{x}}_{k+1} \boxplus \delta{\check{\mathbf{x}}}_{k+1}, {^{C}{\mathbf{p}}}_{s} ,{^G\mathbf{P}_{s}}) }{\partial  \delta{\check{\mathbf{x}}}_{k+1}} |_{\delta{\check{\mathbf{x}}}_{k+1} = \mathbf{0}} \nonumber \\
	{\boldsymbol{\Sigma}_{\boldsymbol{\beta}_{s}}} &= \boldsymbol{\Sigma}_{\mathbf{n}_{\mathbf{p}_{s}}} + {\mathbf{F}_{{{\mathbf{P}}}_{s}}} \boldsymbol{\Sigma}_{\mathbf{{P}}_s} {\mathbf{F}_{{{\mathbf{P}}}_{s}}}^T \label{eq_def_Hc} \\
	{\mathbf{F}_{{{\mathbf{P}}}_{s}}} &=   
	 \dfrac{ \partial  \mathbf{r}_c(\check{\mathbf{x}}_{k+1}, {^{C}{\mathbf{p}}}_{s}, {^G\mathbf{P}_{s}})  }{\partial  {^{G}{\mathbf{P}}}_{s} } \nonumber
\end{align}
The detailed computation of $\mathbf{H}^{{c}}_{s}$ and ${\mathbf{F}_{{{\mathbf{P}}}_{s}}}$ is given in appendix. \ref{sect_app_HC_Fp_beta}

    \subsection{Update of error-state iterated Kalman filter}\label{sec:update}

    Combining the prior distribution (\ref{eq:prior_delta}), the posterior distribution due to LiDAR measurement (\ref{eq_lidar_meas}) and the posterior distribution due to visual measurement (\ref{eq_visual_meas}), we obtain the maximum a posterior (MAP) estimation of $\delta \check{\mathbf{x}}_{k+1}$:
    
\begin{small}
       \begin{equation}
       \begin{split}
           \hspace{-0.5cm}\mathop{\min}_{\delta \check{\mathbf{x}}_{k+1}} &\left(  \left\| \check{\mathbf{x}}_{k+1} \boxminus \hat{\mathbf{x}}_{k+1}  + \boldsymbol{\mathcal{H}} \delta{\check{\mathbf{x}}}_{k+1}   \right\|_{\boldsymbol{\Sigma}_{\delta \hat{\mathbf{x}}_{k+1}}^{-1}} \right. \nonumber \\
        +&\sum\nolimits_{j=1}^{m_l} {  \left\| \mathbf{r}_l(\check{\mathbf{x}}_{k+1}, {^{L}{\mathbf{p}}}_{j})  + \mathbf{H}^{l}_{j} \delta{\check{\mathbf{x}}}_{k+1} \right\|^2_{\boldsymbol{\Sigma}_{\boldsymbol{\alpha}_{j}}^{-1}} }\nonumber \\
		+ & \left.\sum\nolimits_{s=1}^{m_c}{ \left\| \mathbf{r}_c(\check{\mathbf{x}}_{k+1},  {^{C}\mathbf{p}_s}, {^G\mathbf{P}}_{s}) + \mathbf{H}^{{c}}_{s} \delta{\check{\mathbf{x}}}_{k+1} \right\|^2_{\boldsymbol{\Sigma}_{\boldsymbol{\beta}_{s}}^{-1}} } \right)
       \end{split}
        \label{eq_optimial_map}
       \end{equation}
\end{small}
	where $\left\| \mathbf{x} \right\|_{\boldsymbol{\Sigma}}^2 = \mathbf{x} \boldsymbol{\Sigma} \mathbf{x}^T  $. Notice that the measurements of LiDAR and camera may not appear at the same time instant (see Fig. \ref{fig_esikf}), therefore $m_l$ (or $m_c$) could be zero in the above optimization. Denote
	\begin{small}
	\begin{equation}
		\begin{split}
		    \mathbf{H}^T &= 
		\begin{bmatrix}
			{\mathbf{H}^{l}_{1}} ,\dots, {\mathbf{H}^{l}_{m_l}} ,
			{\mathbf{H}^{{c}}_{1}}^T ,\dots, {\mathbf{H}^{{c}}_{m_c}}
		\end{bmatrix}^T\\
		\mathbf{R} &= 
		\text{diag}(
		\begin{matrix}
			\boldsymbol{\Sigma}_{\boldsymbol{\alpha}_{1}} , \dots ,\boldsymbol{\Sigma}_{\boldsymbol{\alpha}_{m_l}} , \boldsymbol{\Sigma}_{\boldsymbol{\beta}_{1}} , \dots ,\boldsymbol{\Sigma}_{\boldsymbol{\beta}_{m_c}} 
		\end{matrix}
		)\\
		\check{\mathbf{z}}_{k+1}^T &= 
		\left[
			 \mathbf{r}_l(\check{\mathbf{x}}_{k+1}, {^{L}{\mathbf{p}}}_{1}),\dots,  
		 \mathbf{r}_l(\check{\mathbf{x}}_{k+1}, {^{L}{\mathbf{p}}}_{m_l}), \right. \\
		 & \quad  \left. \mathbf{r}_c(\check{\mathbf{x}}_{k+1},  {^{C}\mathbf{p}_1}, {^G\mathbf{P}}_{1}),\dots,
		 \mathbf{r}_c(\check{\mathbf{x}}_{k+1},  {^{C}\mathbf{p}_{m_c}}, {^G\mathbf{P}}_{{m_c}})
		\right] \\
		\mathbf{P} &= 
			\left(\boldsymbol{\mathcal{H}}\right)^{-1} \boldsymbol{\Sigma}_{\delta \hat{\mathbf{x}}_{k+1}} {\left(\boldsymbol{\mathcal{H}}\right)}^{-T}
		\end{split}
	\end{equation}
	\end{small}

	 Following \cite{xu2020fast}, we have the Kalman gain computed as:
	\begin{align}
			\mathbf{K} = \left( \mathbf{H}^T\mathbf{R}^{-1}\mathbf{H} + \mathbf{P}^{-1} \right)^{-1}\mathbf{H}^T\mathbf{R}^{-1}
	\end{align}

    Then we can update the state estimate as:
    \begin{small}
    \begin{align}
        \check{\mathbf{x}}_{k+1} =& \check{\mathbf{x}}_{k+1}\boxplus \left( -\mathbf{K}\check{\mathbf{z}}_{k+1} - \left( \mathbf{I}- \mathbf{KH} \right) \left( \boldsymbol{\mathcal{H}} \right)^{-1} \left( \check{\mathbf{x}}_{k+1} \boxminus \hat{\mathbf{x}}_{k+1} \right) \right) \nonumber 
    \end{align}
	\end{small}
	The above process (Section \ref{sec:lidar_meas} to Section \ref{sec:update}) is iterated until convergence (i.e., the update is smaller than a given threshold). The converged state estimate is then used to (1) project points in the the new LiDAR frame to the world frame and append them to the existing point cloud map; (2) triangulate new visual landmarks of the current frame if it is a keyframe; (3) serve as the starting point of the propagation in Section \ref{eq:Propagation} for the next cycle:
	$$
		\hat{\mathbf{x}}_{k+1} = \check{\mathbf{x}}_{k+1}, \hspace{0.2cm}
		\hat{\boldsymbol{\Sigma}}_{\delta \bar{\mathbf{x}}_{k+1}} =  \left( \mathbf{I}- \mathbf{KH} \right)\check{\boldsymbol{\Sigma}}_{\delta \mathbf{x}_{k+1}}
	$$
	
	\section{Factor graph optimization}\label{sec:fact_graph}

	As mentioned in Section \ref{sec:update}, untracked visual landmarks in the newly added keyframe are triangulated to create new visual landmarks. This triangulation is usually of low precision due to keyframe pose estimation error. To further improve the quality of visual landmarks, keyframe poses, and simultaneously calibrate the time offset between the camera and LiDAR-IMU subsystem, we leverage a factor graph optimization for optimizing the camera-poses and the visual landmarks within a sliding window of image keyframes. 
	\begin{figure}[h]
		\centering
		\includegraphics[width=1.0\linewidth]{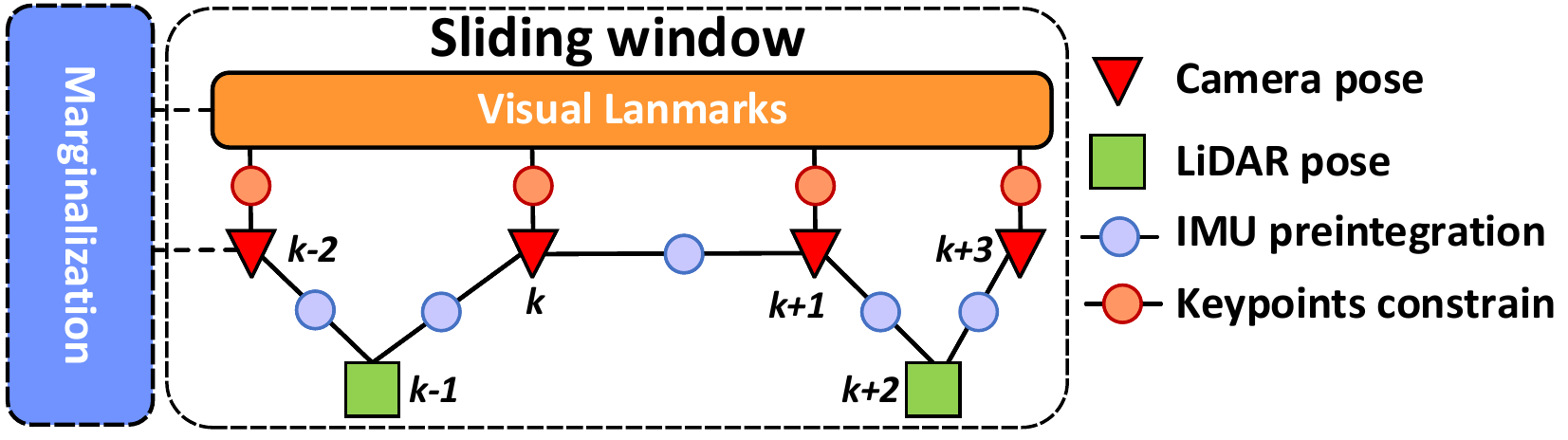}
		\caption{Our factor graph optimization.}
		\label{fig_ssd}
	\end{figure}
	
	Our factor graph optimization is similar to VINS-Mono \cite{qin2018vins}, but further incorporates pose constraints due to LiDAR measurements as shown in Fig. \ref{fig_ssd}. Constraints due to IMU preintegration are also included to connect the LiDAR factor with camera factor. To keep the back-end optimization light-weight, the LiDAR poses in the pose graph are fixed and the LiDAR raw point measurements are not engaged in the pose graph optimization.

	 \section{Experiments and Results}
\subsection{Our device for data sampling}
\begin{figure}[t]
	\centering
	\includegraphics[width=1.0\linewidth]{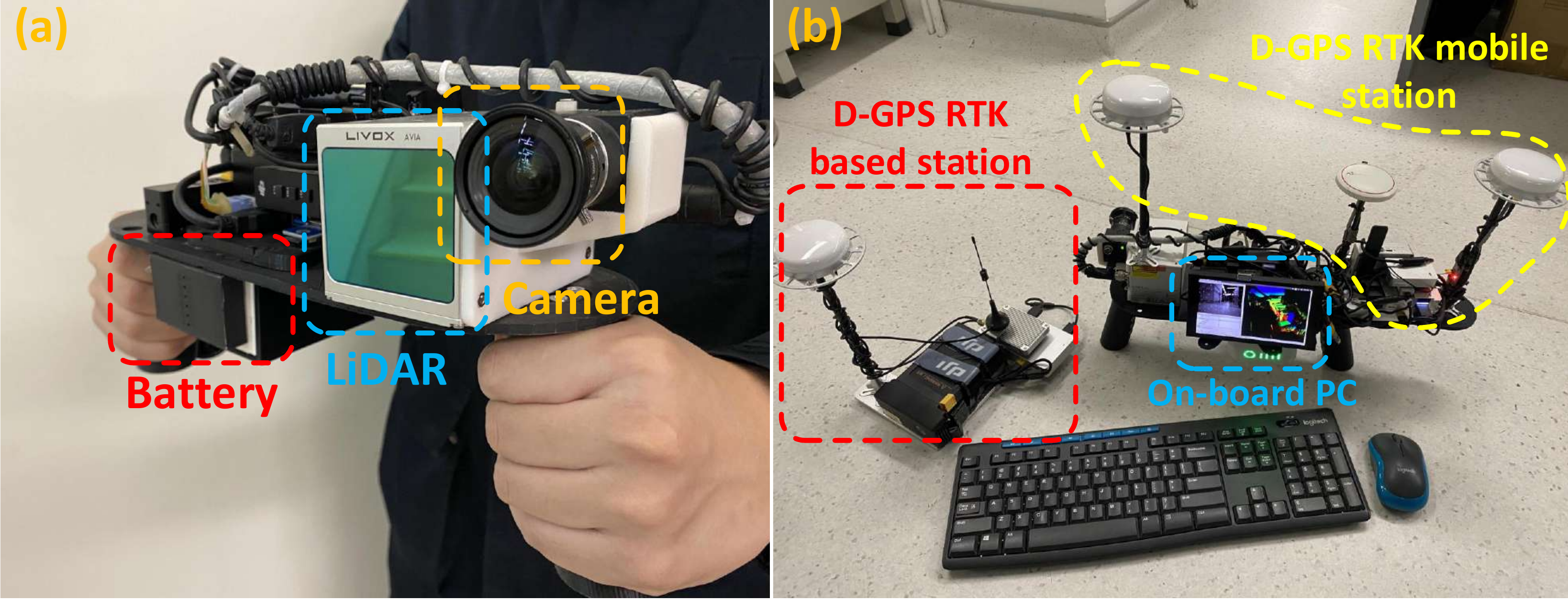}
	\caption{Our handheld device for data sampling,  (a) shows our minimum system, with a total weight $2.09$ Kg; (b) a D-GPS RTK system is used to evaluate the accuracy.}
	\vspace{-1.6cm}
	\label{fig_handheld_device}
\end{figure}

Our handheld device for data sampling is shown in Fig. \ref{fig_handheld_device} (a), which includes the power supply unit,  the onboard DJI \textit{manifold-2c}\footnote{\url{https://www.dji.com/manifold-2}} computation platform (equipped with  an \textit{Intel i7-8550u} CPU and 8 GB RAM), a global shutter camera,  and a \textit{LiVOX AVIA}\footnote{\url{https://www.livoxtech.com/avia}} LiDAR. The FoV of the camera is $82.9^\circ \times 66.5 ^\circ$ while the FoV of LiDAR is $70.4^\circ \times 77.2 ^\circ$. For quantitatively evaluating the precision of our algorithm (our experiment in Section. \ref{sect_experiment_4}), we install a differential-GPS (D-GPS) real-time kinematic (RTK) system\footnote{\url{https://www.dji.com/d-rtk}} on our device, shown in Fig. \ref{fig_handheld_device} (b).
 
\begin{figure}[h]
	\centering
	\includegraphics[width=1.0\linewidth]{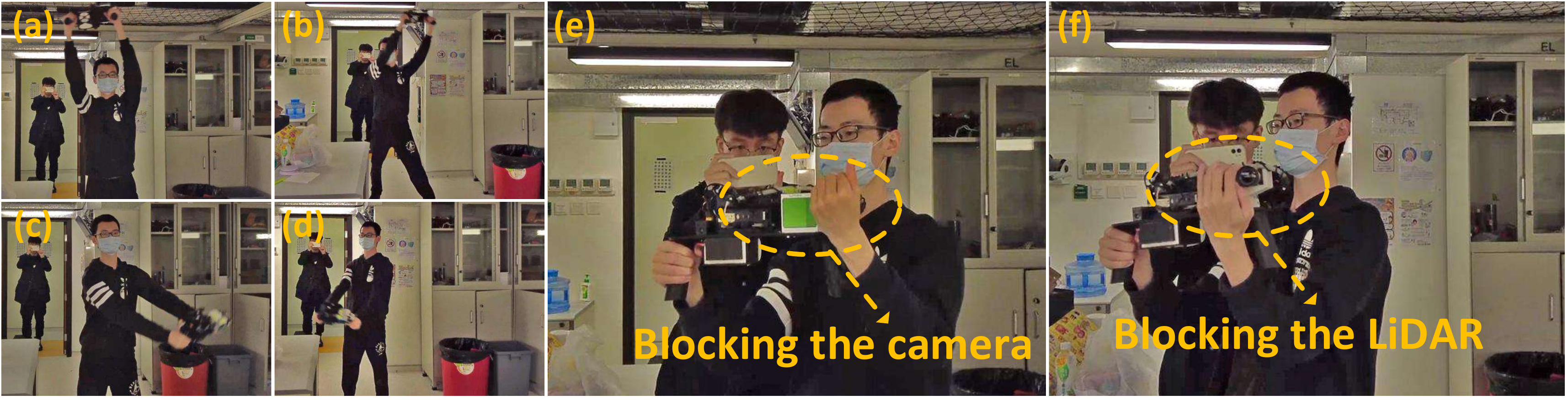}
	\caption{We evaluate the robustness of our algorithm under scenarios with the aggressive motion (sequence in a$\sim$d) and sensor failure by intentionally blocking the camera (e) and LiDAR sensor (f).}
	\label{fig_experi_1_block}
	\includegraphics[width=1.0\linewidth]{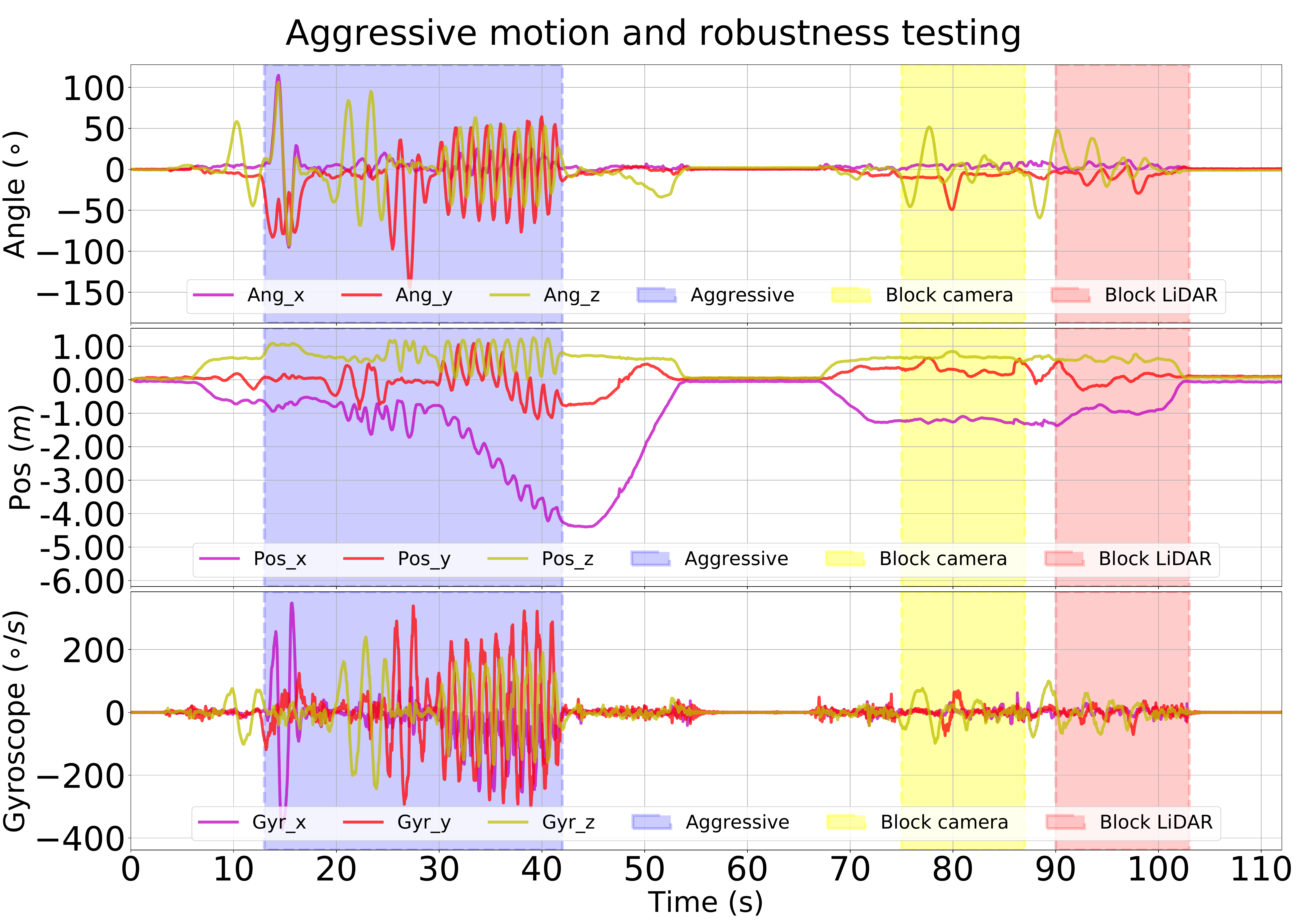}
	\caption{Our estimated pose and the raw gyroscope reading of Experiment-1. The shaded area in blue, yellow and red represent different phases of aggressive motion, camera-failure and LiDAR-failure, respectively.}
	\label{fig_experi_1_curve}
	\vspace{-1cm}
\end{figure}

\begin{figure*}[t]
	\centering
	\includegraphics[width=2.0\columnwidth]{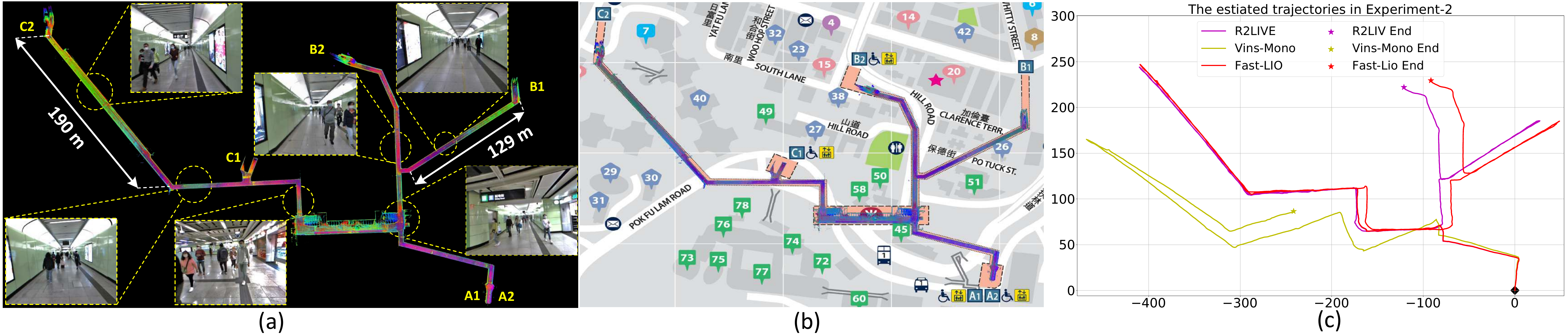}
	\caption{We evaluate our algorithm in a Hong Kong MTR station consisting of cluttered lobby and very long narrow tunnels, as shown in (a). The tunnel is up to $190$ meters long and is filled with moving pedestrians, making it extremely challenging for both LiDAR-based and camera-based SLAM methods. (b): the map built by our system is well aligned with the street map of MTR station. (c) Trajectory comparison among our system {``R2LIVE"}, the LiDAR-inertial system {``Fast-LIO"} , and visual-inertial system {``VINS-Mono"} and  (our). The starting point is marked with $\MyDiamond$ while the ending point of each trajectory is marked with $\bigstar$. {``VINS-Mono"} stopped at middle way due to the failure of feature tracking. }
	\label{fig_subway}
	\vspace{-0.6cm}
\end{figure*}
\subsection{Experiment-1: Robustness evaluation with aggressive motion and sensor failure}\label{sect_experi_1}

In this experiment, we evaluate the robustness of our algorithm under the scenario with aggressive motion (see Fig. \ref{fig_experi_1_block} (a$\sim$d)), in which the maximum angular speed reaches up to $300^\circ /s$ (see the gyroscope readings in Fig. \ref{fig_experi_1_curve}). In addition, we also simulate drastic sensor failures by intentionally blocking the camera (see Fig. \ref{fig_experi_1_block}(e))  and LiDAR
(see Fig. \ref{fig_experi_1_block}(f)). 

The results of our estimated attitude and position are shown in Fig. \ref{fig_experi_1_curve}, where we use shade the different testing phases with different colors. As shown in Fig. \ref{fig_experi_1_curve}, we can see that our estimated trajectory can tightly track the actual motion even in severe rotation and translation. Even when the camera or LiDAR provides no measurements, the estimated trajectory is sill very smooth and exhibits no noticeable degradation. These results show that our algorithm is robust enough to endure with the case with aggressive motion or even with the failure of sensors. We refer readers to the accompanying video on \todo{XXX} showing details of the experiment in practice.
\vspace{-0.5cm}
\subsection{Experiment-2: Robustness evaluation in a narrow tunnel-like environments with moving objects}
In this experiment, we challenge one of the most difficult scenarios in the scope of camera-based and LiDAR-based SLAM, a MTR station, the \textit{HKU station}\footnote{\url{https://en.wikipedia.org/wiki/HKU_station}}. It is a typical narrow tunnel-like environment (see the left figure of Fig. \ref{fig_subway}(a)), with the maximum reaches $190$ meters. Moreover, there are many moving pedestrians frequently showing in the sensor views. All these factors make the SLAM extremely challenging: 1) the long tunnel structure significantly reduces the geometric features for LiDAR-based SLAM, especially when walking along the tunnel or turning the LiDAR around at one end of the tunnel; 2) The highly repeated visual feature (pattern on the floor) causes the error in matching and tracking the visual features in visual SLAM; 3) The many moving pedestrians could cause outliers to the already few point- and visual-features.

Despite these challenges, our system is robust enough to survive in this scene. In Fig. \ref{fig_subway} (b), we align our point cloud data with the \textit{HKU station street map}\footnote{\url{https://www.mtr.com.hk/archive/en/services/maps/hku.pdf}}, and find them match tightly. This demonstrates that our localization is of high robustness and accuracy. Moreover, we plot our estimated trajectory together with that of Vins-Mono\footnote{\url{https://github.com/HKUST-Aerial-Robotics/VINS-Mono}} and Fast-LIO\footnote{\url{https://github.com/hku-mars/FAST_LIO/}} in Fig. \ref{fig_subway} (c), where we can see that our method achieves the best overall performance in this experiment. 
\begin{figure}[h]
	\centering
	{
	    \vspace{-0.5cm}
		\includegraphics[width=0.8\linewidth]{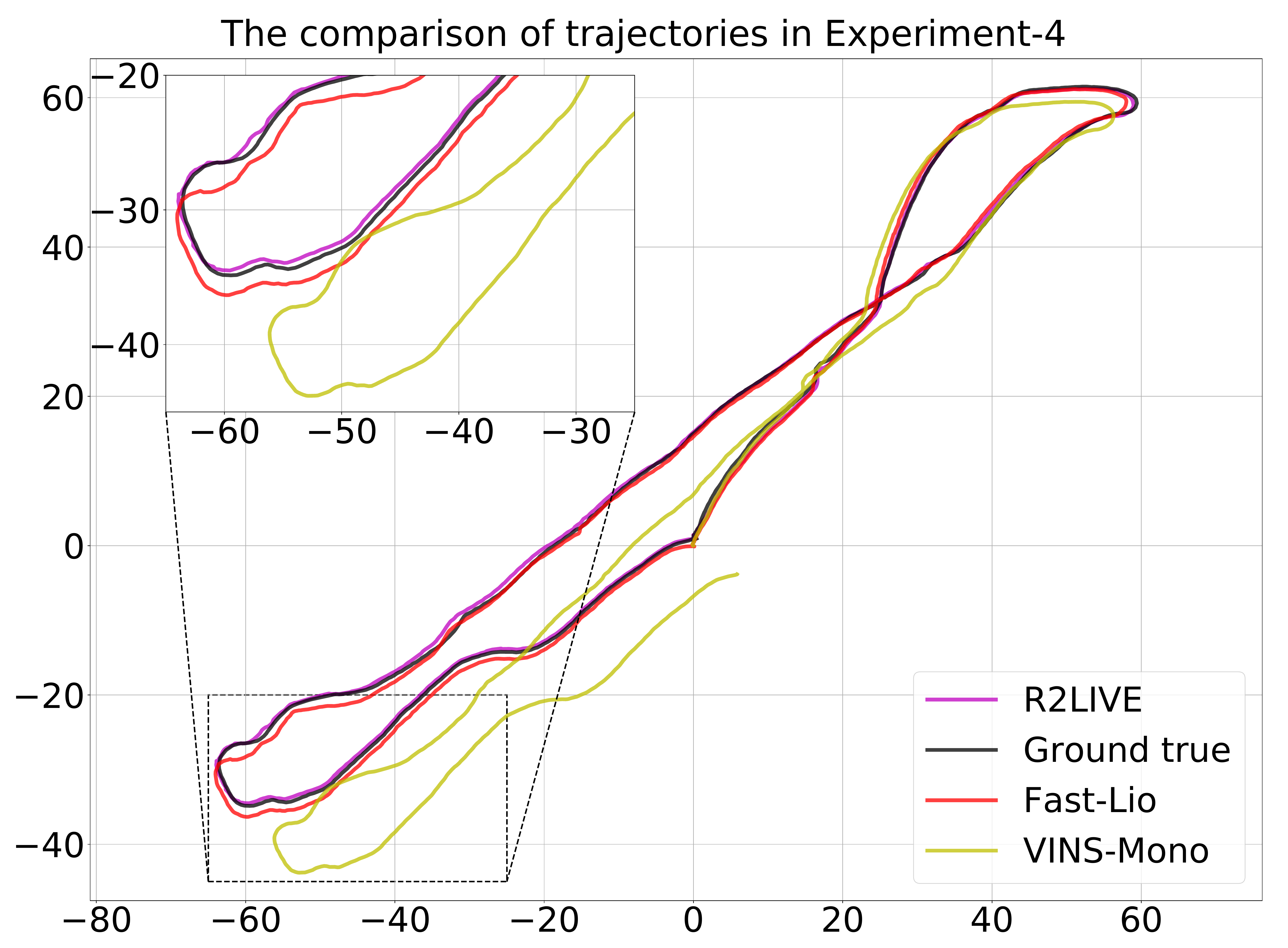}
		\includegraphics[width=0.8\linewidth]{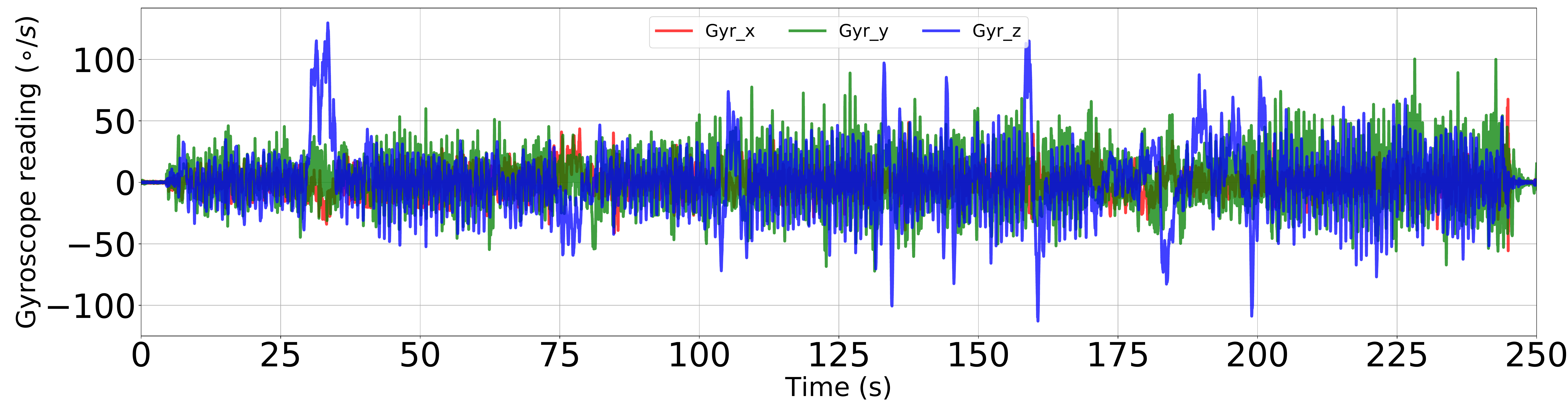}
		\caption{The upper figure plots the different trajectories in experiments-4, while the lower figure shows the raw gyroscope reading in the experiment. }
		\label{Fig_eval_rtk}
		\vspace{-0.5cm}
	}
\end{figure}
\begin{figure}[h]
	\centering
	\includegraphics[width=1.0\linewidth]{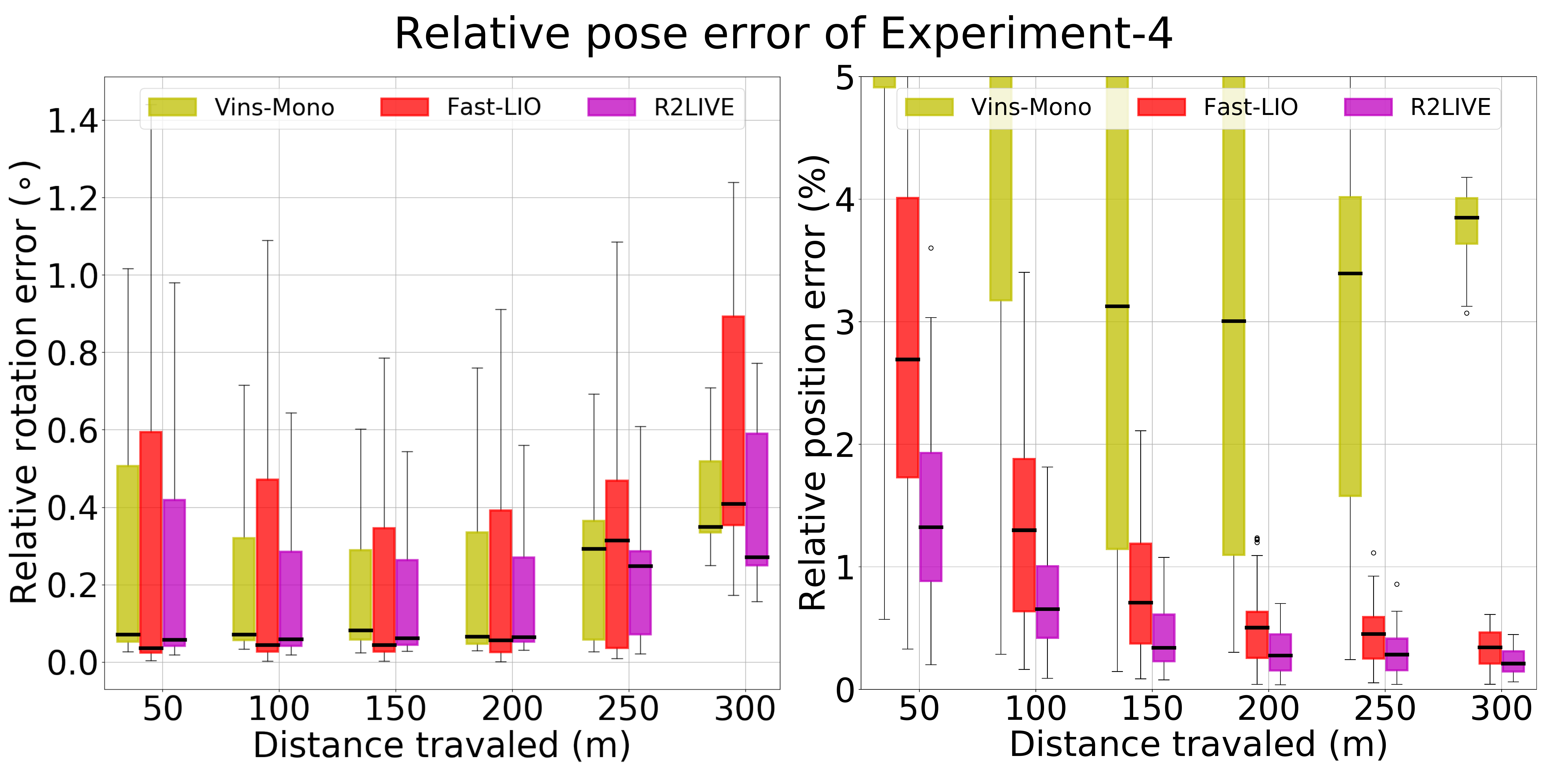}
	\caption{The relative pose error in experiments-4, for the sequence of $300$ meters, the median of pose error of ``{{Vins-Mono}}", ``{Fast-LIO}" and ``{R2LIVE}" are ($0.35^\circ$,$3.84 \%$), ($0.41^\circ$,$0.34 \%$) and ($\mathbf{0.27}^\circ$,$\mathbf{0.21}\%$) respectively.}
	\label{fig_dgps_rpe}
	\vspace{-0.5cm}
\end{figure}
\begin{figure*}[t]
	\centering
	\centering
	\includegraphics[width=1.0\linewidth]{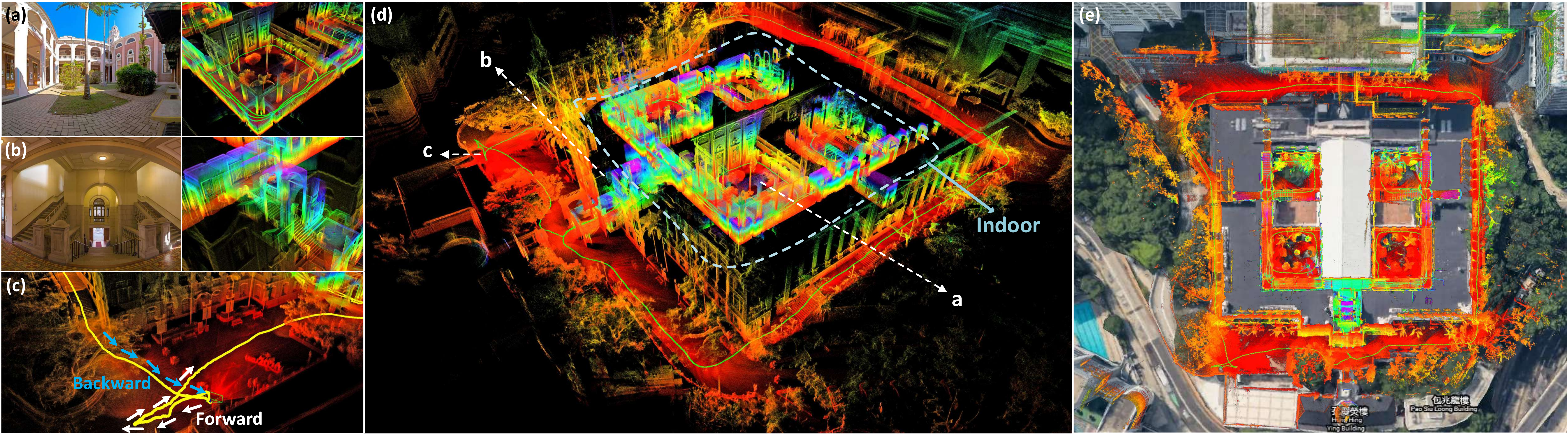}
	\caption{The reconstructed 3D maps in Experiment-3 are shown in (d), and the detail point cloud with the correspondence panorama images are shown in (a) and (b). (c) show that our algorithm can close the loop itself (returning the starting point) without any additional processing (e.g. loop closure). In (e), we merge our maps with the satellite image to further examine the accuracy of our system.}
	\vspace{-0.5cm}
	\label{fig_hku_main_building}
\end{figure*}
\subsection{Experiment-3: High precision maps building in large-scale indoor \& outdoor urban environment}
In this experiment, we show that our proposed method is accurate enough to reconstruct a dense 3D, high precision, large-scale indoor-outdoor map of urban environments.  We collect data of the \textit{HKU main building} exterior and interior. The real-time reconstructed 3D maps is shown in Fig. \ref{fig_hku_main_building}, in which the clear and high-quality point cloud demonstrates that our proposed method is of high accuracy. What worth to be mentioned is, without any additional processing (i.e. loop closure), our algorithm can still close the loop itself (see Fig. \ref{fig_hku_main_building} (c)) after traversing $876$ meters, which demonstrates that our proposed method is extremely low drift. Finally, we merge our maps together with the satellite image and find them match tightly (see Fig. \ref{fig_hku_main_building} (e)). 

\subsection{Experiment-4: Quantitative evaluation of precision using D-GPS RTK}\label{sect_experiment_4}

In this experiment, we quantitatively evaluate the precision of Vins-Mono (IMU+Camera), Fast-LIO (IMU+LiDAR) and our algorithm by comparing their estimated trajectory with the ground truth trajectory (see the upper figure of Fig. \ref{Fig_eval_rtk}) obtained from a real-time differential-GPS (D-GPS) Kinematic (RTK). The data has the maximum angular velocity reaching $130^\circ/s$ (see the gyroscope reading in Fig. \ref{Fig_eval_rtk}). We calculate translational and rotational errors for all possible subsequences of length (50,...,300) meters, with the relative pose error (RPE) among these methods shown in Fig. \ref{fig_dgps_rpe}.

\subsection{Run time analysis}
The average time consumption in experiment 1$\sim$4 are listed in TABLE. \ref{tab2}, which demonstrates that R$^2$LIVE can achieve real-time on both the desktop PC and embedded computation platform. Noticed that the factor graph optimization is run on a separate thread and therefore is allowed to run at a lower rate.

\section{Conclusion}
 In this letter, we propose a robust, real-time LiDAR-inertial-visual fusion framework based on a high-rate filter-based odometry and factor graph optimization. We fuse the measurements of LiDAR, inertial, and camera sensors within an error-state iterated Kalman filter and use the factor graph optimization to refine the local map that in a sliding window of image keyframes and visual landmarks. Our system was tested in large-scale, indoor-outdoor, and narrow tunnel-like environments, challenging sensor failure and aggressive motion. In all the tests, our method achieves a high level of accuracy and robustness in localization and mapping.

\begin{table}[]
	\begin{scriptsize}
	\begin{tabular}{|c|c|c|c|c|}
		\hline
		PC/on-board & Exp-1 (ms)         & Exp-2 (ms)      & Exp-3 (ms)  &  Exp-4 (ms) \\ \hline
		LI-Odom       & 8.81 / 15.98 & 11.54 / 25.30 & 14.91 / 30.64 & 10.92 / 24.37 \\ \hline
		VI-Odom       &  7.84 / 13.92 &  8.84 / 19.10  & 9.57 / 19.56 &  8.99 / 20.16  \\ \hline
		FG-OPM      &  26.10 / 45.35 & 30.20 / 65.25 & 29.25 / 58.08 & 27.98 / 60.43 \\ \hline
	\end{tabular}
	\caption{The average running time of R2LIVE in Experiment-1$\sim$4 (Exp-1$\sim$4) on desktop PC (with Intel i7-9700K  CPU and 32GB RAM) and on-board computed (with Intel i7-8550u CPU and 8GB RAM). The items "LI-Odom", "VI-Odom" and "FG-OPM" are the average time consumption of LiDAR-IMU filter-based odometry, Visual-Inertial filter-based odometry, and factor graph optimization, respectively.
	 }
	\label{tab2}
	\end{scriptsize}
	\vspace{-1cm}
\end{table}

    \bibliography{r2liv}
    \vspace{-0.3cm}
    \appendix
\subsection{Computation of $\mathbf{F}_{\delta{\hat{\mathbf{x}}}}$ and $\mathbf{F}_{\mathbf{w}}$ } \label{sect_app_fx_fw}
\begin{scriptsize}
	\begin{align}
		&\mathbf{F}_{\delta{\hat{\mathbf{x}}}} = \left. \dfrac{ \partial \left( \delta{\hat{\mathbf{x}}}_{i+1} \right)  }{\partial\delta{\hat{\mathbf{x}}_i}} \right|_{\delta{\hat{\mathbf{x}}_i} = \mathbf{0}, \mathbf{w}_i = \mathbf{0}} \nonumber\\
		=&
		\begin{bmatrix}
			\mathtt{Exp}( -\hat{\boldsymbol{\omega}}_i\Delta t )  & \mathbf{0} & \mathbf{0} & \mathbf{0}  &  \mathbf{0} & -{\mathbf{J}_r(\hat{\boldsymbol{\omega}}_i\Delta t )}^T & \mathbf{0}    \\
			\mathbf{0} & \mathbf{I}  & \mathbf{0} & \mathbf{0} & \mathbf{I}\Delta t &  \mathbf{0} & \mathbf{0} \\
			\mathbf{0} & \mathbf{0} &  \mathbf{I} & \mathbf{0} & \mathbf{0} & \mathbf{0} & \mathbf{0}   \\
			\mathbf{0} & \mathbf{0} &  \mathbf{0} & \mathbf{I} & \mathbf{0} & \mathbf{0} & \mathbf{0}   \\
			-^G\hat{\mathbf{R}}_{I_i} [\hat{\mathbf{a}}_{i}]_\times \Delta t & \mathbf{0} & \mathbf{0} & \mathbf{0} & \mathbf{I}   & \mathbf{0} &  -^G\hat{\mathbf{R}}_{I_i} \Delta t \\
			\mathbf{0} & \mathbf{0} & \mathbf{0} & \mathbf{0} & \mathbf{0} &  \mathbf{I} & \mathbf{0} \\
			\mathbf{0} & \mathbf{0} & \mathbf{0} & \mathbf{0} & \mathbf{0} &  \mathbf{0} & \mathbf{I}  
		\end{bmatrix} \nonumber \\
		&\mathbf{F}_{{\mathbf{w}}} = \left. \dfrac{ \partial \left( \delta{\hat{\mathbf{x}}}_{i+1} \right)  }{\partial{\mathbf{w}_i}} \right|_{\delta{\hat{\mathbf{x}}_i} = \mathbf{0}, \mathbf{w}_i = \mathbf{0}} \nonumber \\
		=& 
		\begin{bmatrix}
			-{\mathbf{J}_r(\hat{\boldsymbol{\omega}}_i\Delta t )}^T & \mathbf{0} & \mathbf{0} & \mathbf{0} \\
			\mathbf{0}& \mathbf{0}& \mathbf{0}& \mathbf{0}\\
			\mathbf{0}& \mathbf{0}& \mathbf{0}& \mathbf{0}\\
			\mathbf{0}& \mathbf{0}& \mathbf{0}& \mathbf{0}\\
			\mathbf{0}& -^G\hat{\mathbf{R}}_{I_i} \Delta t & \mathbf{0}& \mathbf{0}\\
			\mathbf{0}& \mathbf{0}& \mathbf{I}\Delta t& \mathbf{0}\\
			\mathbf{0}& \mathbf{0}& \mathbf{0}& \mathbf{I}\Delta t\\
		\end{bmatrix}
		\nonumber
	\end{align}
\end{scriptsize}
where $\hat{\boldsymbol{\omega}}_{i} = \boldsymbol{\omega}_{m_i} - \mathbf{b}_{\mathbf{g}_i} $ , $\hat{\mathbf{a}}_{i} = \mathbf{a}_{m_i} - \mathbf{b}_{\mathbf{a}_i} $, and $\mathbf{J}_r(\cdot)$ are called the \textit{right Jacobian matrix} of $SO(3)$:
\begin{small}
$$
\mathbf{J}_r(\mathbf{r}) =  \mathbf{I} - \dfrac{1-\cos||\mathbf{r}||}{||\mathbf{r}||^2}\left[\mathbf{r}\right]_\times + \dfrac{||\mathbf{r}||-\sin(||\mathbf{r}||)}{||\mathbf{r}||^3}\left[\mathbf{r}\right]_\times^2, \mathbf{r}\in \mathbb{R}^3
$$
\end{small}
For the detailed derivation of $\mathbf{F}_{\check{\delta{\mathbf{x}}}}$ and $\mathbf{F}_{\mathbf{w}}$, please refer to the  Section. B of our supplementary material.
\vspace{-0.3cm}
\subsection[The computation of ${\mathcal{H}}$]{The computation of $\boldsymbol{\mathcal{H}}$} \label{sect_app_Hx}
\begin{align}
	\boldsymbol{\mathcal{H}} &=  \dfrac{ \left( \check{\mathbf{x}}_{k+1} \boxplus \delta \check{\mathbf{x}}_{k+1}  \right) \boxminus \hat{\mathbf{x}}_{k+1} }{\partial  \delta \check{\mathbf{x}}_{k+1}} |_{ \delta \check{\mathbf{x}}_{k+1}  = \mathbf{0}} \nonumber \\
	&= 
	\begin{bmatrix}
		\mathbf{A} & \mathbf{0} & \mathbf{0} &  \mathbf{0} & \mathbf{0}_{3\times 9} \\
		\mathbf{0} & \mathbf{I} & \mathbf{0} & \mathbf{0} & \mathbf{0}_{3\times 9}  \\
		\mathbf{0} & \mathbf{0} & \mathbf{B} &  \mathbf{0} &\mathbf{0}_{3\times 9}\\
		\mathbf{0} & \mathbf{0} & \mathbf{0} &  \mathbf{I} &\mathbf{0}_{3\times 9} \\
		\mathbf{0} & \mathbf{0} & \mathbf{0} &  \mathbf{0} &\mathbf{I}_{9\times 9} \\
	\end{bmatrix} \nonumber
\end{align}
where the $3\times 3$ matrix $\mathbf{A} = \mathbf{J}_r^{-1} ( \mathtt{Log}({{^G\hat{\mathbf{R}}_{I_{k+1}}}^T}{^G\check{\mathbf{R}}_{I_{k+1}}} ) )$ and $
\mathbf{B} = \mathbf{J}_r^{-1} ( \mathtt{Log}({{^I\hat{\mathbf{R}}_{C_{k+1}}}^T}{^I\check{\mathbf{R}}_{C_{k+1}}} ) ) $. $\mathbf{J}_r^{-1}(\cdot)$ are called the \textit{inverse right Jacobian matrix} of $SO(3)$:
\begin{small}
	\begin{align}
	\mathbf{J}_r^{-1}(\mathbf{r}) = & \mathbf{I} + \dfrac{1}{2}\left[\mathbf{r}\right]_\times+  \left(\dfrac{1}{||\mathbf{r}||^2} - \dfrac{1+\cos(||\mathbf{r}||)}{2||\mathbf{r}||\sin(||\mathbf{r}||)}\right)	\left[\mathbf{r}\right]_\times^2 
	\end{align}
\end{small}

\vspace{-0.6cm}
\subsection{The computation of $\mathbf{H}^{{l}}_{j}$} \label{sect_app_HL_alpha}
\begin{align}
	\mathbf{H}^{{l}}_{j} = \mathbf{u}_j^T
	\begin{bmatrix}
		- {^G\check{\mathbf{R}}_{I_{k+1}}} \left[ \mathbf{P_a}\right]_\times & \mathbf{I}_{3\times 3} & \mathbf{0}_{3\times 15}
	\end{bmatrix} \nonumber 
\end{align}
where $\mathbf{P_a} =  {^I\mathbf{R}}_{L}{^{L}{\mathbf{p}}_j}  + {^{I}\mathbf{p}_L} $.

For the detailed derivation of $\mathbf{H}^{{l}}_{j}$, please refer to the Section. D of our supplementary material.

\subsection{The computation of $\mathbf{H}^{{c}}_{s}$ and ${\mathbf{F}_{{{\mathbf{P}}}_{s}}}$} \label{sect_app_HC_Fp_beta}
\vspace{-0.4cm}
\begin{align}
	&\mathbf{H}^{{c}}_{s} 
	= - \mathbf{F_A} \cdot \mathbf{F_B} \nonumber \\
	&\mathbf{F}_{{{\mathbf{P}}}_{s}} =- \mathbf{F_A} \cdot \mathbf{F_C} \nonumber
\end{align}
with:
\begin{align}
	\mathbf{F_A} & = \dfrac{1}{{^{C}{P}_s}_z}
	\begin{bmatrix}
		f_x & 0 & -f_x \dfrac{{^{C}{P}_s}_x}{{^{C}{P}_s}_z} \\
		0 & f_y & -f_y \dfrac{{^{C}{P}_s}_y}{{^{C}{P}_s}_z}
	\end{bmatrix} \\
	\mathbf{F_B} & =
	\begin{bmatrix}
		\mathbf{M_A} & \mathbf{M_B} & \mathbf{M_C} & -\mathbf{I} & \mathbf{0}_{3\times 12}
	\end{bmatrix} \nonumber \\
	\mathbf{F_C} & = \left({^G\check{\mathbf{R}}_{I_{k+1}}}{^I\check{\mathbf{R}}_{C}} \right)^T 
\end{align}
where $f_x$ and $f_y$ are the focal length, $c_x$ and $c_y$ are the principal point offsets in image plane, and the $3\times 3$ matrix $\mathbf{M_A}$, $\mathbf{M_B}$ and  $\mathbf{M_C}$ are:
\begin{align}
	&\mathbf{M_A} = \left({^I\check{\mathbf{R}}_{C}}\right)^T\left[\left({^G\hat{\mathbf{R}}_{I_{k+1}}}\right)^T{{^G}{\mathbf{P}}_s}\right]_\times \nonumber \\
	&\mathbf{M_B} = - \left({^I\hat{\mathbf{R}}_{C}}\right)^T \nonumber\\
	&\mathbf{M_C} = \left[ \left( {^G\hat{\mathbf{R}}_{I_{k+1}}}{^I\hat{\mathbf{R}}_{C}}\right)^T{{^G}{\mathbf{P}}_s}  \right]_\times -\left[\left({^I\hat{\mathbf{R}}_{C}}\right)^T{{^G}\hat{\mathbf{p}}^i_{I_{k+1}}}\right]_\times \nonumber
\end{align}

For the detailed derivation of $\mathbf{H}^{{c}}_{s}$ and ${\mathbf{F}_{{{\mathbf{P}}}_{s}}}$, please refer to  Section. E of our supplementary material.
\clearpage

\setcounter{equation}{0}
\setcounter{figure}{0}
\setcounter{table}{0}
\setcounter{page}{1}
\setcounter{section}{1}
\setcounter{section}{0}%
\setcounter{subsection}{0}%
\setcounter{subsubsection}{0}%
\setcounter{paragraph}{0}%

\begin{center}
	\textbf{\large Supplementary Material: R$^2$LIVE: A Robust, Real-time, LiDAR-Inertial-Visual tightly-coupled state Estimator and mapping}
\end{center}

\renewcommand{\thesection}{S\arabic{section}}
\renewcommand{\theequation}{S\arabic{equation}}
\renewcommand{\thefigure}{S\arabic{figure}}


\subsection{Perturbation on $SO(3)$}\label{sect_supply_perturbation}
In this appendix, we will use the following approximation of perturbation $\delta\mathbf{r}\rightarrow \mathbf{0}$ on $SO(3)$ \cite{sola2017quaternion, barfoot2017state}:
\begin{align}
	\mathtt{Exp}(\mathbf{r}  + \delta\mathbf{r}) &\approx 
	\mathtt{Exp}(\mathbf{r}) \mathtt{Exp}(\mathbf{J}_r(\mathbf{r})\delta\mathbf{r}) \nonumber \\
	\mathtt{Exp}(\mathbf{r})\mathtt{Exp}(\delta\mathbf{r}) &\approx \mathtt{Exp}(\mathbf{r} + \mathbf{J}_r^{-1}(\mathbf{r}) \delta\mathbf{r}  ) \nonumber \\
	\mathbf{R}\cdot\mathtt{Exp}(\delta\mathbf{r})\cdot\mathbf{u} &\approx \mathbf{R}\left( \mathbf{I}+ \left[\delta\mathbf{r}\right]_\times\right)\mathbf{u} = \mathbf{R}\mathbf{u}- \mathbf{R}\left[\mathbf{u}\right]_\times\delta \mathbf{r} \nonumber 
\end{align}
where $\mathbf{u}\in \mathbb{R}^{3}$ and we use $\left[\cdot\right]_\times$ denote the skew-symmetric matrix of vector $(\cdot)$; $\mathbf{J}_r(\mathbf{r})$ and $\mathbf{J}_r^{-1}(\mathbf{r})$ are called the \textit{right Jacobian} and the \textit{inverse right Jacobian} of $SO(3)$, respectively.
\begin{align}
	\mathbf{J}_r(\mathbf{r}) = & \mathbf{I} - \dfrac{1-\cos||\mathbf{r}||}{||\mathbf{r}||^2}\left[\mathbf{r}\right]_\times + \dfrac{||\mathbf{r}||-\sin(||\mathbf{r}||)}{||\mathbf{r}||^3}\left[\mathbf{r}\right]_\times^2  \nonumber \\
	\mathbf{J}_r^{-1}(\mathbf{r}) = & \mathbf{I} + \dfrac{1}{2}\left[\mathbf{r}\right]_\times+  \left(\dfrac{1}{||\mathbf{r}||^2} - \dfrac{1+\cos(||\mathbf{r}||)}{2||\mathbf{r}||\sin(||\mathbf{r}||)}\right)	\left[\mathbf{r}\right]_\times^2 \nonumber
\end{align}

\subsection{Computation of $\mathbf{F}_{\delta{\mathbf{x}}}$ and $\mathbf{F}_{\mathbf{w}}$ } 

Combing (\ref{eq_def_delta_x_i}) and (\ref{eq_delta_x_k_plus_1}) , we have:
\begin{align}
&\quad\quad\delta{\hat{\mathbf{x}}}_{i+1} = {\mathbf{x}}_{i+1} \boxminus \hat{\mathbf{x}}_{i+1}  \nonumber \\
& =\Large(\mathbf{x}_{i} \boxplus \left( \Delta t \cdot \mathbf{f}({\mathbf{x}}_i, \mathbf{u}_i, \mathbf{w}_i) \right)  \Large) \boxminus \left( \hat{\mathbf{x}}_{i}\boxplus
\left( \Delta t \cdot \mathbf{f}(\hat{\mathbf{x}}_i, \mathbf{u}_i, \mathbf{0}) \right)\right)  \nonumber
\end{align}
	\vspace{-0.4cm}
	\begin{scriptsize}
	$$
	\hspace{-0.2cm}
	=\begin{bmatrix}
	\mathbf{Log}\left( \left( ^G\hat{\mathbf{R}}_{I_i} \mathtt{Exp}\left( {\hat{\boldsymbol{\omega}}}_{i} \Delta t \right)\right)^{T} \cdot \left( ^G\hat{\mathbf{R}}_{I_i}\mathtt{Exp}\left({^G\delta \mathbf{r}_{I_i}} \right)  \mathtt{Exp}\left( {\boldsymbol{\omega}}_{i} \Delta t \right)   \right)	\right) \\
	{^G\delta\mathbf{p}_{I_i}} +   {^G\delta\mathbf{v}_{i}} \Delta t + \dfrac{1}{2}{\mathbf{a}}_i \Delta t^2 - \dfrac{1}{2}{\hat{\mathbf{a}}}_i \Delta t^2\\
	{^{I}\delta\mathbf{r}_{C_i}} \\
	{^{{I}}\delta\mathbf{p}_{C_i}} \\
	{^G\delta\mathbf{v}_{i}} + \left( ^G\hat{\mathbf{R}}_{I_i}\mathtt{Exp}\left({^G\delta \mathbf{r}_{I_i}} \right)     \right){\mathbf{a}}_i \Delta t - {^G\hat{\mathbf{R}}_{I_i}} \hat{\mathbf{a}}_i \Delta t \\
	\delta\mathbf{b}_{g_i} + \mathbf{n}_{\mathbf{bg}_i}  \\
	\delta\mathbf{a}_{g_i} + \mathbf{n}_{\mathbf{ba}_i} \\
	\end{bmatrix} \nonumber
	$$

\end{scriptsize}
with:
\begin{align}
	\hat{\boldsymbol{\omega}}_{i} = \boldsymbol{\omega}_{m_i} - \mathbf{b}_{\mathbf{g}_i}, &\hspace{0.2cm}
 {\boldsymbol{\omega}}_{i}  = \hat{\boldsymbol{\omega}}_{i} - \delta\mathbf{b}_{\mathbf{g}_i} -\mathbf{n}_{\mathbf{g}_i}  \\
 ~\hat{\mathbf{a}}_{i} = \mathbf{a}_{m_i} - \mathbf{b}_{\mathbf{a}_i} , &\hspace{0.2cm} {\mathbf{a}}_{i} =   \hat{\mathbf{a}}_{i} - \delta\mathbf{b}_{\mathbf{a}_i} -\mathbf{n}_{\mathbf{a}_i}
\end{align}

And we have the following simplification and approximation form Section. A. 
\begin{footnotesize}
\begin{align}
\hspace{-0.2cm} &\mathtt{Log}\left( \left( ^G\hat{\mathbf{R}}_{I_i} \mathtt{Exp}\left( {\hat{\boldsymbol{\omega}}}_{i} \Delta t \right)\right)^{T} \cdot \left( ^G\hat{\mathbf{R}}_{I_i}\mathtt{Exp}\left({^G\delta \mathbf{r}_{I_i}} \right)  \mathtt{Exp}\left( {\boldsymbol{\omega}}_{i} \Delta t \right)   \right)	\right) \nonumber \\
=&	\mathtt{Log}\left( \mathtt{Exp}\left( \hat{\boldsymbol{\omega}}_{i} \Delta t \right)^{T} \cdot \left( \mathtt{Exp}\left({^G\delta \mathbf{r}_{I_i}} \right) \cdot \mathtt{Exp}\left( {\boldsymbol{\omega}}_{i} \Delta t \right)   \right)	\right) \nonumber \\
\approx& \mathtt{Log}\left( \mathtt{Exp}\left( \hat{\boldsymbol{\omega}}_{i} \Delta t \right)^{T}\mathtt{Exp}\left({^G\delta \mathbf{r}_{I_i}} \right)  \mathtt{Exp}\left( \hat{\boldsymbol{\omega}}_{i} \Delta t \right)\cdot \right. \nonumber \\
&\hspace{0.8cm}\left. \mathtt{Exp}\left( -\mathbf{J}_r( \hat{\boldsymbol{\omega}}_{i}\Delta t  )\left( \delta\mathbf{b}_{g_i} + \mathbf{n}_{\mathbf{g}_i} \right) \right)  \right) \nonumber \\
\approx& \mathtt{Exp}\left( \hat{\boldsymbol{\omega}}_{i}\Delta t \right) \cdot  {^G\delta \mathbf{r}_{I_i}}  -{\mathbf{J}_r(\hat{\boldsymbol{\omega}}_i\Delta t )}^T\delta\mathbf{b}_{\mathbf{g}_i} -{\mathbf{J}_r(\hat{\boldsymbol{\omega}}_i\Delta t )}^T\mathbf{n}_{\mathbf{g}_i} \nonumber \\
&\left( ^G\mathbf{R}_{I_i}\mathtt{Exp}\left({^G\delta \mathbf{r}_{I_i}} \right)    \right){\mathbf{a}}_i \Delta t \nonumber\\
\approx& \left( ^G\mathbf{R}_{I_i} \left( \mathbf{I} + [^G\delta \mathbf{r}_{I_i}]_\times \right)     \right) \left( \hat{\mathbf{a}}_i - \delta\mathbf{b}_{\mathbf{a}_i} -\mathbf{n}_{\mathbf{a}_i}\right) \Delta t \nonumber\\
\approx& ^G\mathbf{R}_{I_i} \hat{\mathbf{a}}_i \Delta t - ^G\mathbf{R}_{I_i} \delta\mathbf{b}_{\mathbf{a}_i} \Delta t - ^G\mathbf{R}_{I_i} \mathbf{n}_{\mathbf{a}_i} \Delta t- {^G\mathbf{R}_{I_i}} \left[ \hat{\mathbf{a}}_i \right]_\times  {^G\delta \mathbf{r}_{I_i}} \nonumber
\end{align}
\end{footnotesize}
To conclude, we have the computation of $\mathbf{F}_{\delta{\mathbf{x}}}$ and $\mathbf{F}_{\mathbf{w}}$ as follow:
\begin{scriptsize}
\begin{align}
	&\mathbf{F}_{\delta{\hat{\mathbf{x}}}} = \left. \dfrac{ \partial \left( \delta{\hat{\mathbf{x}}}_{i+1} \right)  }{\partial\delta{\hat{\mathbf{x}}_i}} \right|_{\delta{\hat{\mathbf{x}}_i} = \mathbf{0}, \mathbf{w}_i = \mathbf{0}} \nonumber\\
	=&
	\begin{bmatrix}
		\mathtt{Exp}( -\hat{\boldsymbol{\omega}}_i\Delta t )  & \mathbf{0} & \mathbf{0} & \mathbf{0}  &  \mathbf{0} & -{\mathbf{J}_r(\hat{\boldsymbol{\omega}}_i\Delta t )}^T & \mathbf{0}    \\
		\mathbf{0} & \mathbf{I}  & \mathbf{0} & \mathbf{0} & \mathbf{I}\Delta t &  \mathbf{0} & \mathbf{0} \\
		\mathbf{0} & \mathbf{0} &  \mathbf{I} & \mathbf{0} & \mathbf{0} & \mathbf{0} & \mathbf{0}   \\
		\mathbf{0} & \mathbf{0} &  \mathbf{0} & \mathbf{I} & \mathbf{0} & \mathbf{0} & \mathbf{0}   \\
		-^G\hat{\mathbf{R}}_{I_i} [\hat{\mathbf{a}}_{i}]_\times \Delta t & \mathbf{0} & \mathbf{0} & \mathbf{0} & \mathbf{I}   & \mathbf{0} &  -^G\hat{\mathbf{R}}_{I_i} \Delta t \\
		\mathbf{0} & \mathbf{0} & \mathbf{0} & \mathbf{0} & \mathbf{0} &  \mathbf{I} & \mathbf{0} \\
		\mathbf{0} & \mathbf{0} & \mathbf{0} & \mathbf{0} & \mathbf{0} &  \mathbf{0} & \mathbf{I}  
	\end{bmatrix} \nonumber
\end{align}
\end{scriptsize}
\begin{small}
\begin{align}
	&\mathbf{F}_{{\mathbf{w}}} = \left. \dfrac{ \partial \left( \delta{\hat{\mathbf{x}}}_{i+1} \right)  }{\partial{\mathbf{w}_i}} \right|_{\delta{\hat{\mathbf{x}}_i} = \mathbf{0}, \mathbf{w}_i = \mathbf{0}} \nonumber \\
	=& 
	\begin{bmatrix}
	-{\mathbf{J}_r(\hat{\boldsymbol{\omega}}_i\Delta t )}^T & \mathbf{0} & \mathbf{0} & \mathbf{0} \\
	 \mathbf{0}& \mathbf{0}& \mathbf{0}& \mathbf{0}\\
	 \mathbf{0}& \mathbf{0}& \mathbf{0}& \mathbf{0}\\
	 \mathbf{0}& \mathbf{0}& \mathbf{0}& \mathbf{0}\\
	 \mathbf{0}& -^G\hat{\mathbf{R}}_{I_i} \Delta t & \mathbf{0}& \mathbf{0}\\
	 \mathbf{0}& \mathbf{0}& \mathbf{I}\Delta t& \mathbf{0}\\
	 \mathbf{0}& \mathbf{0}& \mathbf{0}& \mathbf{I}\Delta t\\
	\end{bmatrix}
	\nonumber
\end{align}
\end{small}
\vspace{-0.5cm}
\subsection[The computation of ${\mathcal{H}}$]{The computation of $\boldsymbol{\mathcal{H}}$} 
Recalling (\ref{eq_def_HL_gamma_alpha}), we have:
\begin{align}
	\boldsymbol{\mathcal{H}} &=  \dfrac{ \left( \check{\mathbf{x}}_{k+1} \boxplus \delta \check{\mathbf{x}}_{k+1}  \right) \boxminus \hat{\mathbf{x}}_{k+1} }{\partial  \delta \check{\mathbf{x}}_{k+1}} |_{ \delta \check{\mathbf{x}}_{k+1}  = \mathbf{0}} \nonumber \\
	&=
	\begin{bmatrix}
	\mathbf{A} & \mathbf{0} & \mathbf{0} &  \mathbf{0} & \mathbf{0}_{3\times 9} \\
		\mathbf{0} & \mathbf{I} & \mathbf{0} & \mathbf{0} & \mathbf{0}_{3\times 9}  \\
		\mathbf{0} & \mathbf{0} & \mathbf{B} &  \mathbf{0} &\mathbf{0}_{3\times 9}\\
		\mathbf{0} & \mathbf{0} & \mathbf{0} &  \mathbf{I} &\mathbf{0}_{3\times 9} \\
		\mathbf{0} & \mathbf{0} & \mathbf{0} &  \mathbf{0} &\mathbf{I}_{9\times 9} \\
	\end{bmatrix}\nonumber
\end{align}
with the $3\times 3$ matrix $\mathbf{A} = \mathbf{J}_r^{-1} ( \mathtt{Log}({{^G\hat{\mathbf{R}}_{I_{k+1}}}^T}{^G\check{\mathbf{R}}_{I_{k+1}}} ) )$ and $
\mathbf{B} = \mathbf{J}_r^{-1} ( \mathtt{Log}({{^I\hat{\mathbf{R}}_{C_{k+1}}}^T}{^I\check{\mathbf{R}}_{C_{k+1}}} ) ) $.

\subsection{The computation of $\mathbf{H}^{{l}}_{j}$} 
Recalling (\ref{eq_def_rl_x_kplus1}) and (\ref{eq_def_HL_gamma_alpha}), we have:
\begin{align}
	 &\mathbf{r}_l(\check{\mathbf{x}}_{k+1} \boxplus \delta\check{\mathbf{x}}_{k+1}, {^{L}{\mathbf{p}}}_{j})  =\mathbf{u}_j^T\left(  {^G\check{\mathbf{p}}_{I_{k+1}}} + {^G\delta\check{\mathbf{p}}_{I_{k+1}}} - \right. \nonumber \\
	 & {{\mathbf{q}}_j } + {^G\check{\mathbf{R}}_{I_{k+1}}}\mathtt{Exp}({^G\check{\delta\mathbf{r}}_{I_{k+1}}})\left. ( ^I\mathbf{R}_{L}{^{L}{\mathbf{p}}_j}  + {^{I}\mathbf{p}_L} )  \right) \label{eq_rl_x_delta_x_p_alpha} 
\end{align}
And with the small perturbation approximation, we get: 
\begin{small}
\begin{align}
	&{^G\check{\mathbf{R}}_{I_{k+1}}}\mathtt{Exp}({^G\delta\check{\mathbf{r}}_{I_{k+1}}}) \mathbf{P_a}  \nonumber \\
	\approx & {^G\check{\mathbf{R}}_{I_{k+1}}}\left( \mathbf{I} + \left[ {^G\delta\check{\mathbf{r}}_{I_{k+1}}} \right]_\times\right) \mathbf{P_a} \nonumber \\
	= & {^G\check{\mathbf{R}}_{I_{k+1}}} \mathbf{P_a} - {^G\check{\mathbf{R}}_{I_{k+1}}} \left[ \mathbf{P_a}\right]_\times {^G\delta\check{\mathbf{r}}_{I_{k+1}}} \label{eq_G_R_I_kplus1_exp}
\end{align}
\end{small}
where $\mathbf{P_a} =  {^I\mathbf{R}}_{L}{^{L}{\mathbf{p}}_j}  + {^{I}\mathbf{p}_L} $. Combining (\ref{eq_rl_x_delta_x_p_alpha}) and (\ref{eq_G_R_I_kplus1_exp}) together we can obtain:
\begin{align}
	\mathbf{H}^{{l}}_{j} = \mathbf{u}_j^T
	\begin{bmatrix}
		- {^G\check{\mathbf{R}}_{I_{k+1}}} \left[ \mathbf{P_a}\right]_\times & \mathbf{I}_{3\times 3} & \mathbf{0}_{3\times 15}
	\end{bmatrix} \nonumber 
\end{align}

\subsection{The computation of $\mathbf{H}^{{c}}_{s}$ and ${\mathbf{F}_{{{\mathbf{P}}}_{s}}}$} 

Recalling (\ref{eq_visual_residual}), we have:
$${^C\mathbf{P}_{s}} = \mathbf{P}_{\mathbf{C}}(\check{\mathbf{x}}_{k+1} , {^G\mathbf{P}_{s}}) = \begin{bmatrix}
	{^{C}{P}_s}_x ~ {^{C}{P}_s}_y ~ {^{C}{P}_s}_z
\end{bmatrix}^T  $$
where the function $\mathbf{P}_{\mathbf{C}}(\check{\mathbf{x}}_{k+1} , {^G\mathbf{P}_{s}})$ is:
\begin{align}
	&\mathbf{P}_{\mathbf{C}}(\check{\mathbf{x}}_{k+1} , {^G\mathbf{P}_{s}}) = \left({^G\check{\mathbf{R}}_{I_{k+1}}}{^I\check{\mathbf{R}}_{C_{k+1}}} \right)^T  {{^G}{\mathbf{P}}_s} \\
	& \hspace{2.5cm}	 - \left({^I\check{\mathbf{R}}_{C_{k+1}}}\right)^T  {{^G}\check{\mathbf{p}}_{I_{k+1}}} -{^I\check{\mathbf{p}}_{C_{k+1}}} \label{eq_app_C_def_Pb}
\end{align} 

From  (\ref{eq_def_Hc}), we have:
\begin{small}
	\begin{align}
		 \mathbf{r}_c\left({\check{\mathbf{x}}_{k+1}, {^{C}{\mathbf{p}}}_{s}}, {^G\mathbf{P}_{s}}\right) & = 
		{^{C}}\mathbf{p}_{s} - \boldsymbol{\pi}( {^C\mathbf{P}_{s}} ) \nonumber \\
		\boldsymbol{\pi}( {^C\mathbf{P}_{s}} ) & = \begin{bmatrix}
			f_x\dfrac{ {^{C}{P}_s}_x }{{^{C}{P}_s}_z} + c_x ~~
			f_y\dfrac{ {^{C}{P}_s}_y }{{^{C}{P}_s}_z} + c_y  
		\end{bmatrix}^T \label{eq_app_C_proj}
	\end{align}
\end{small} 
where $f_x$ and $f_y$ are the focal length, $c_x$ and $c_y$ are the principal point offsets in image plane. 

For conveniently, we omit the $(\cdot)|_{\delta{\check{\mathbf{x}}}^i_{k+1} = \mathbf{0}}$ in the following derivation, and we have:
\begin{align}
	&\mathbf{H}^{{c}}_{s} 
	= - \dfrac{ \partial  \boldsymbol{\pi}( {^C\mathbf{P}_{s}} ) }{\partial  {^C\mathbf{P}_{s}} } \cdot
	\dfrac{ \partial  \mathbf{P}_{\mathbf{C}}(\check{\mathbf{x}}_{k+1} \boxplus \delta \check{\mathbf{x}}_{k+1} , {^G\mathbf{P}_{s}})   }{ \partial  \delta{\check{\mathbf{x}}}_{k+1} } \label{eq_subpply_def_Hc} \\
	&\mathbf{F}_{{{\mathbf{P}}}_{s}} =- \dfrac{ \partial  \boldsymbol{\pi}( {^C\mathbf{P}_{s}} ) }{\partial  {^C\mathbf{P}_{s}} }\cdot
	 \dfrac{\partial \mathbf{P}_{\mathbf{C}}(\check{\mathbf{x}}_{k+1} , {^G\mathbf{P}_{s}})}{\partial  {^{G}{\mathbf{P}}}_{s}} \label{eq_subpply_def_Fps} 
\end{align}
where:
\begin{small}
\begin{align}
\dfrac{ \partial  \boldsymbol{\pi}( {^C\mathbf{P}_{s}} ) }{\partial  {^C\mathbf{P}_{s}} } &= \dfrac{1}{{^{C}{P}_s}_z}
	\begin{bmatrix}
	f_x & 0 & -f_x \dfrac{{^{C}{P}_s}_x}{{^{C}{P}_s}_z} \\
	0 & f_y & -f_y \dfrac{{^{C}{P}_s}_y}{{^{C}{P}_s}_z}
	\end{bmatrix} \label{eq_supply_visual_dpi_dpc} \\
	\dfrac{\partial \mathbf{P_b}(\check{\mathbf{x}}_{k+1}  , {^G\mathbf{P}_{s}})}{\partial  {^{G}{\mathbf{P}}}_{s}} &=  \left({^G\check{\mathbf{R}}_{I_{k+1}}}{^I\check{\mathbf{R}}_{C}} \right)^T \label{eq_supply_visual_dpc_dps}
\end{align}
\end{small}
According to Section. A, we have the following approximation of $\mathbf{P}_{\mathbf{C}}(\check{\mathbf{x}}_{k+1} \boxplus \delta \check{\mathbf{x}}_{k+1} , {^G\mathbf{P}_{s}})$:
\begin{footnotesize}
\begin{align}
	 &\hspace{0.2cm}\mathbf{P}_{\mathbf{C}}(\check{\mathbf{x}}_{k+1} \boxplus \delta \check{\mathbf{x}}_{k+1} , {^G\mathbf{P}_{s}}) \nonumber \\
	 & \hspace{-0.2cm}=\left({^G\check{\mathbf{R}}_{I_{k+1}}} 
	 \mathtt{Exp}\left( {^G{\delta \check{\mathbf{r}}_{I_{k+1}}}} \right)
	  {^I\check{\mathbf{R}}_{C_{k+1}}} 
	  \mathtt{Exp}\left( {^I{\delta \check{\mathbf{r}}_{C_{k+1}}}} \right)
	  \right)^T  {{^G}{\mathbf{P}}_s} - \hspace{-0.1cm}{^I\check{\mathbf{p}}_{C}} \nonumber \\
	 &\hspace{-0.2cm}  - {^I{\delta\check{\mathbf{p}}}_{C}} -\left({^I\check{\mathbf{R}}_{C}} \mathtt{Exp}\left( {^I{\delta \check{\mathbf{r}}_{C}}} \right)\right)^T \left(  {{^G}\check{\mathbf{p}}_{I_{k+1}}} + {^G\delta\check{\mathbf{p}}}_{I_{k+1}}  \right)  \nonumber \\
	 &\hspace{-0.2cm} \approx  \mathbf{P_b}(\check{\mathbf{x}}^i_{k+1} , {^G\mathbf{P}_{s}}) + 
	 \left[ \left( {^G\check{\mathbf{R}}_{I_{k+1}}}{^I\check{\mathbf{R}}_{C}}\right)^T{{^G}{\mathbf{P}}_s}  \right]_\times  {^I{\delta \check{\mathbf{r}}_{C}}} \nonumber \\
	 &\hspace{-0.2cm}+\left({^I\check{\mathbf{R}}_{C}}\right)^T\left[\left({^G\check{\mathbf{R}}_{I_{k+1}}}\right)^T{{^G}{\mathbf{P}}_s}\right]_\times {^G{\delta \check{\mathbf{r}}_{I_{k+1}}}} - \left({^I\check{\mathbf{R}}_{C}}\right)^T {^G\delta\check{\mathbf{p}}}_{I_{k+1}} \nonumber \\
	 &\hspace{-0.2cm}-\left[\left({^I\check{\mathbf{R}}_{C}}\right)^T{{^G}\check{\mathbf{p}}_{I_{k+1}}}\right]_\times {^I{\delta\check{\mathbf{r}}}_{C}} - {^I{\delta\check{\mathbf{p}}}_{C}} \nonumber
\end{align}
\end{footnotesize}
With this, we can derive:
\begin{footnotesize}
\begin{align}
	&\dfrac{\partial \mathbf{P}_{\mathbf{C}}(\check{\mathbf{x}}_{k+1} \boxplus \delta \check{\mathbf{x}}_{k+1} , {^G\mathbf{P}_{s}})  }{ \partial  \delta{\check{\mathbf{x}}}_{k+1} } = \begin{bmatrix}
		\mathbf{M_A}~~ \mathbf{M_B} ~~ \mathbf{M_C} ~~ -\mathbf{I} ~~ \mathbf{0}_{3\times 12}
	\end{bmatrix} \label{eq_supply_visual_dpc_dx} \\
	&\mathbf{M_A} = \left({^I\check{\mathbf{R}}_{C}}\right)^T\left[\left({^G\hat{\mathbf{R}}_{I_{k+1}}}\right)^T{{^G}{\mathbf{P}}_s}\right]_\times \nonumber \\
	&\mathbf{M_B} = - \left({^I\hat{\mathbf{R}}_{C}}\right)^T \nonumber\\
	&\mathbf{M_C} = \left[ \left( {^G\hat{\mathbf{R}}_{I_{k+1}}}{^I\hat{\mathbf{R}}_{C}}\right)^T{{^G}{\mathbf{P}}_s}  \right]_\times -\left[\left({^I\hat{\mathbf{R}}_{C}}\right)^T{{^G}\hat{\mathbf{p}}^i_{I_{k+1}}}\right]_\times \nonumber
\end{align}
\end{footnotesize}
Substituting (\ref{eq_supply_visual_dpi_dpc}), (\ref{eq_supply_visual_dpc_dps}) and (\ref{eq_supply_visual_dpc_dx}) into (\ref{eq_subpply_def_Hc}) and (\ref{eq_subpply_def_Fps}), we finish the computation of $\mathbf{H}^{{c}}_{s}$ and ${\mathbf{F}_{{{\mathbf{P}}}_{s}}}$.


\end{document}